%% file: Main.tex
\documentclass[11pt, letterpaper]{article}
\usepackage[margin=1in]{geometry} 
\usepackage{multirow}
\usepackage{subfig}
\usepackage{tikzpeople}
\usepackage{paralist}
\usepackage[square,numbers]{natbib}
\usepackage{authblk}
\usepackage{hyperref}

\usepackage{amsthm}
\usepackage{amsmath}
\usepackage{amssymb}
\usepackage{pifont}
\usepackage{ragged2e}
\usepackage{array}
\usepackage{marvosym}
\usepackage{enumitem}
\usepackage[capitalise,noabbrev]{cleveref}
\newtheorem{theorem}{Theorem}

\newtheorem{lemma}[theorem]{Lemma}
\theoremstyle{definition}
\newtheorem{definition}{Definition}

\newcommand{\cmark}{\ding{51}}%

\AtBeginDocument{%
  \providecommand\BibTeX{{%
    \normalfont B\kern-0.5em{\scshape i\kern-0.25em b}\kern-0.8em\TeX}}}

\begin{document}

\title{Responsible AI for General-Purpose Systems:
Overview, Challenges, and A Path Forward}

\author{Gourab K. Patro}
\author{Himanshi Agrawal}
\author{Himanshu Gharat}
\author{Supriya Panigrahi}
\author{Nim Sherpa}
\author{Vishal Vaddina}
\author{Dagnachew Birru}
\affil{Phi Labs, Quantiphi Inc.}

\maketitle
\begin{abstract}
Modern general-purpose AI systems made using large language and vision models, are capable of performing a range of tasks like writing text articles, generating and debugging codes, querying databases, and translating from one language to another, which has made them quite popular across industries.
However, there are risks like hallucinations, toxicity, and stereotypes in their output that make them untrustworthy.
We review various risks and vulnerabilities of modern general-purpose AI along eight widely accepted responsible AI (RAI) principles (fairness, privacy, explainability, robustness, safety, truthfulness, governance, and sustainability) and compare how they are non-existent or less severe and easily mitigable in traditional task-specific counterparts.
We argue that this is due to the non-deterministically high Degree of Freedom in output (DoFo) of general-purpose AI (unlike the deterministically constant or low DoFo of traditional task-specific AI systems), and there is a need to rethink our approach to RAI for general-purpose AI.
Following this, we derive C\textsuperscript{2}V\textsuperscript{2} (\textit{Control}, \textit{Consistency}, \textit{Value}, \textit{Veracity}) desiderata to meet the RAI requirements for future general-purpose AI systems, and discuss how recent efforts in AI alignment, retrieval-augmented generation, reasoning enhancements, etc. fare along one or more of the desiderata.
We believe that the goal of developing responsible general-purpose AI can be achieved by formally modeling application- or domain-dependent RAI requirements along C\textsuperscript{2}V\textsuperscript{2} dimensions, and taking a system design approach to suitably combine various techniques to meet the desiderata.

\end{abstract}
\input{1_intro}
\input{2_related}
\input{3_pitfalls}
\input{31_fairness}
\input{32_privacy}
\input{33_explainability}
\input{34_robustness}
\input{35_safety}
\input{36_truthfulness}
\input{37_governance}
\input{38_sustainability}
\input{4_wayforward}

\input{5_conclusion}

\bibliographystyle{abbrvnat}
\bibliography{references}

\input{Appendix}
\end{document}

%% file: 1_intro.tex
\section{Introduction}
\label{sec:intro}
In recent years, advances in Generative AI (\textbf{GenAI}) \citep{gozalo2023survey} ---like Large Language Models (\textbf{LLMs}), Diffusion Models (\textbf{DMs}), and Multimodal Large Language Models (\textbf{MLLMs})---have been quite transformative, rapidly reshaping industries, research, and even our everyday interactions. 
Due to their adaptability and versatility, these GenAI models have been highly popular across industries and are being adopted for tasks like interpreting and responding to human language queries, generating cohesive texts, writing code, generating realistic images and other media content, and automating a number of enterprise-grade tasks;
Thus, the AI systems built on top of these GenAI models are rightly called General-Purpose AI (\textbf{GPAI}).
However, with their tremendous potential comes significant responsibility to avert adverse effects on humans and society.
Recent studies have revealed that these systems sometimes generate plausible-sounding yet factually incorrect or fabricated information, often called as hallucinations \citep{ji2023survey};
additionally, they have also been found generating toxic content \citep{wen2023unveiling} and stereotypical responses \citep{gallegos2024bias}.
Such instances raise concerns regarding the usage of GPAI in domains like healthcare, finance, recruitment etc.
Thus, GPAI systems should be made to operate in a manner that is ethical, transparent, and aligned with societal values, consequently demanding rapid evolution and advancements in the Responsible AI (\textbf{RAI}) discourse in the era of GPAI. 
\begin{figure*}[t]
\includegraphics[width=0.95\textwidth]{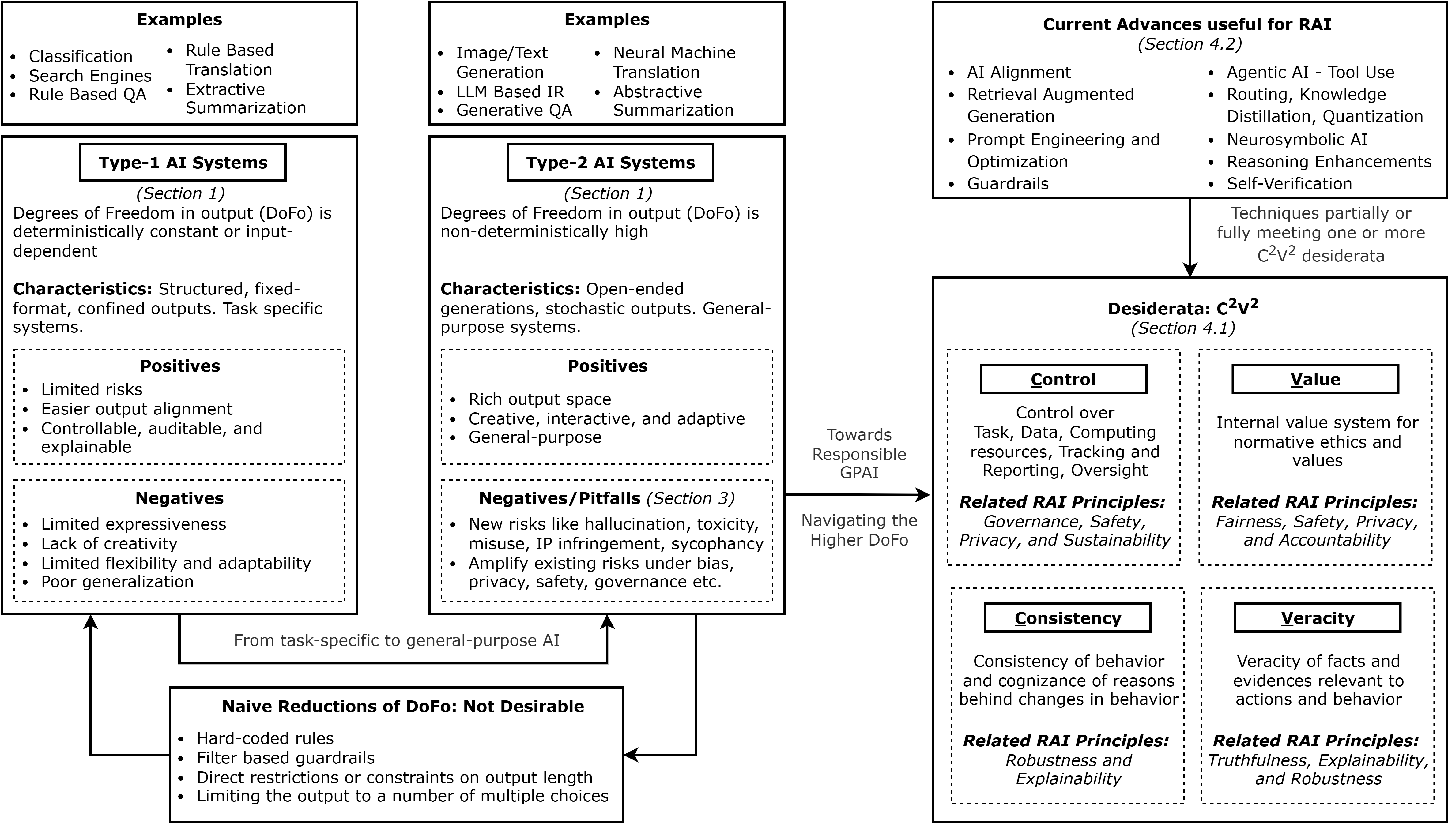}
\caption{\small \bf This figure presents a comprehensive view of our position on RAI through a system's perspective informed by the notion of Degrees of Freedom in Output (DoFo). We begin by distinguishing Type-1 and Type-2 AI systems in (Section \ref{sec:intro}) based on their DoFo. This distinction shows how the shift toward general-purpose AI amplifies existing RAI risks and also introduces new ones, which we elucidate in (Section \ref{sec:pitfalls}). Next, we illustrate two divergent approaches to navigate the higher DoFo in Type-2 systems: (i) naive reductions that suppress generative potential by reverting to rigid, Type-1-like constraints that is of course not desirable, and (ii) a principled path forward (Section \ref{sec:path_forward}) where the navigation is guided by C\textsuperscript{2}V\textsuperscript{2} desiderata and supported by emerging techniques that meet or improve along the one or more of C\textsuperscript{2}V\textsuperscript{2} goals, and offer a pathway toward Responsible General-Purpose AI.}
\label{fig:summary} 
\end{figure*} 
~\\{\bf Types of AI Systems and Their Degrees of Freedom in Output:}
RAI has long been studied for traditional task-specific AI (classification, regression, heuristic and rule-based AI, symbolic AI, etc.), and significant breakthroughs have been achieved in mitigating RAI risks (even with mathematical guarantees) \citep{kaur2022trustworthy,cheng2021socially}.
Thus, one might wonder whether we can simply transfer our understandings, modeling, and methodologies of RAI in task-specific AI to modern GPAI.
In this context, taking a system's perspective (considering AI as a system that given certain types of input along with additional support data or tools, it either gives an output or performs a certain task), we would like to point out that traditional task-specific AI systems and modern GPAI systems have a fundamental difference;
it is their degree of freedom in output (\textbf{DoFo}), the maximum number of values that may vary independently in the output space given an input instance.
We find that the DoFo of task-specific AI systems is deterministically low and constant, or depends on input size;
For example AI systems designed for classification, regression, rule-based translation and chatbots have fixed DoFo, whereas AI systems for extractive summarization and index-based search have DoFo dependent on input sizes (theoretical analyses are given in the Appendix).
In contrast, modern GPAI systems being used for general information retrieval (\textbf{IR}), machine translations (\textbf{MT}), generative question-answering (\textbf{QA}), abstractive summarization, etc. can have non-deterministically high DoFo due to their ability to generate (usually uncountably infinite) textual responses having orthogonally different semantics. 
Thus, we classify task-specific AI systems with deterministically constant or input-dependent DoFo as \textbf{Type-1 AI} systems and modern GPAI with non-deterministically high DoFo as \textbf{Type-2 AI} systems.

Type-2 AI systems, with very high DoFo, bring in many new capabilities and can perform a range of tasks (with no/low explicit training).
However, they also end up being vulnerable to newer and more severe RAI risks than their Type-1 counterparts;
We detail these differences in RAI risks (along widely accepted RAI principles: fairness, privacy, explainability, robustness, safety, governance, and sustainability) for Type-1 and Type-2 AI systems in Section \ref{sec:pitfalls}.
Thus, we argue that Type-2 AI systems need to be guided or steered through the very high DoFo to meet the essential RAI requirements.
Based on this, we propose the main desiderata for approaching responsible GPAI: \textbf{C\textsuperscript{2}V\textsuperscript{2}} (\textit{Control}, \textit{Consistency}, \textit{Value}, \textit{Veracity});
In Section \ref{subsec:desiderata}, we discuss C\textsuperscript{2}V\textsuperscript{2} desiderata in detail and how they can lead to responsible GPAI.
Some efforts like AI alignment \citep{wang2024comprehensive}, retrieval-augmented generation \citep{lewis2020retrieval} and neurosymbolic AI \citep{garcez2023neurosymbolic} meet or improve GPAI along the C\textsuperscript{2}V\textsuperscript{2} desiderata (Section \ref{subsec:efforts_rgpai}).
However, we believe that the goal of developing responsible GPAI systems can be achieved by formally modeling application- or domain-dependent RAI requirements along C\textsuperscript{2}V\textsuperscript{2}, and taking a system design approach (Section \ref{subsec:sys_design}) to suitably combine various techniques to meet the desiderata.

Moreover, our goals in this paper are to discuss how modern Type-2 (GPAI) AI is at higher risk ---of RAI violations---than traditional Type-1 (task-specific) AI, and lay a broader agenda for future work on developing responsible GPAI.
In summary, we make the following contributions:
\textit{(i)} Taking a system's perspective, we formally differentiate traditional task-specific AI systems (Type-1 AI) from modern GPAI systems (Type-2 AI) using their DoFo;
\textit{(ii)} We discuss (with examples) how Type-2 AI with non-deterministically higher DoFo is more vulnerable to newer and more severe RAI-related risks than their Type-1 counterparts;
\textit{(iii)} Finally, we map the risks in Type-2 AI to a set of desiderata (C\textsuperscript{2}V\textsuperscript{2}) providing a path forward towards responsible GPAI.
%
~\\\textbf{Outline}: Figure \ref{fig:summary} depicts a big picture of the paper and summarizes our position on RAI in the era of GPAI.
Section \ref{sec:related} has a brief overview of the RAI literature.
%
Section \ref{sec:pitfalls} provides a detailed comparison of RAI risks in Type-1 and Type-2 AI systems,
Section \ref{sec:path_forward} presents an agenda and a set of desiderata for responsible GPAI, and 
Section \ref{sec:conclusion} finally concludes.

%% file: 2_related.tex
\section{Overview of Related Work}
\label{sec:related}
{\label{350277}}

\subsection{Responsible AI Principles and Practices:}
\label{subsec:rai_overview}
Responsible AI (RAI) \citep{herrera2025responsible} refers to the design, development, deployment, and use of AI systems that are ethical, trustworthy, and aligned with human values and legal norms, ensuring it benefits individuals and society.
Central to RAI are of core principles discussed across global guidelines and frameworks \citep{unesco_aiethics, oecd2024principles}. Some widely accepted principles include: \textbf{\textit{Fairness}} (equitable treatment across user groups), \textbf{\textit{Privacy}} (protecting sensitive data and system integrity), \textbf{\textit{Explainability}} (understanding how decisions are made), \textbf{\textit{Robustness}} (accurate and consistent performance under adverse conditions), \textbf{\textit{Safety}} (preventing harmful outputs), \textbf{\textit{Truthfulness}} (providing factual and verifiable outputs), \textbf{\textit{Governance}} (accountability, auditability, and oversight), and \textbf{\textit{Sustainability}} (environmental, social, and economic well-being in AI lifecycle).
RAI practices \citep{morley2020initial} provide mechanisms to operationalize these principles. Key practices include \textit{documentation templates} \citep{mitchell2019model}, \textit{impact assessments} \citep{yeung2021guidance}, \textit{human-in-the-loop mechanisms} \citep{wu2022survey}, \textit{fairness evaluations and bias mitigations} \citep{guo2024bias}, \textit{designing and testing AI safety} \citep{chen2024trustworthy}, \textit{explanation methods} \citep{nandkishore2024transparency}, and \textit{red teaming} \citep{chua2024ai}.
These practices are encouraged or mandated by \textit{\textbf{guidelines}}: IEEE Ethics for Autonomous and Intelligent Systems \citep{chatila2019ieee}, 
High-Level Expert Group guidelines for Trustworthy AI \citep{ai2019high}, 
UNESCO Principles on Ethical AI \citep{unesco_aiethics}, \textit{\textbf{laws and regulations}} \citep{eu2024regulation}, \textit{\textbf{frameworks}}: NIST AI Risk Management Framework \citep{ai2023artificial}, \textit{\textbf{standards}}: ISO/IEC 42001:2023 AI Management Systems \citep{iso42001} etc.; which underscores increasing need and consensus about adoption of RAI. 
 
In this paper, we analyze how these principles apply to different categories of AI systems. We show how different risks can be reasonably addressed by existing approaches in traditional AI systems (Type-1 AI), and also highlight how newer generative systems (Type-2 AI) introduce new challenges, amplify existing risks, and demand novel strategies. We also propose methodologies that help bridge the gap to meet RAI requirements in Type-2 AI systems.
\subsection{Contextual Uniqueness of the Paper:}\label{subsec:novelty}
We differentiate our work from a number of related survey, review or position papers on \textit{(i)} generative AI (GenAI) models, \textit{(ii)} specific RAI principles, and \textit{(iii)} governance and regulatory frameworks for RAI.

Our work is evidently different from many review and survey works on LLMs \citep{wei2022emergent,chang2024survey,liang2024survey}, diffusion models \citep{yang2023diffusion,croitoru2023diffusion}, and other GenAI 
models and applications \citep{gozalo2023survey,li2024large,zhu2023large}, that focus mainly on the technical advancements and innovative applications of GenAI models. 
In contrast, we qualitatively investigate ---through a structured interdisciplinary literature review---the additional risks and difficulties in achieving RAI objectives for new GenAI systems.

Several survey, review, and position papers have been written focusing on specific RAI principles:
\textbf{\textit{fairness}} \citep{mehrabi2021survey,finocchiaro2021bridging,gallegos2024bias,parraga2025fairness,bartl2025gender},
\textbf{\textit{privacy}} 
\citep{liu2021machine,rigaki2023survey,yao2024survey,das2025security,zhao2025visual},
\textbf{\textit{explainability}} \citep{zhang2018visual,carvalho2019machine,burkart2021survey,das2020opportunities,zhao2024explainability},
\textbf{\textit{robustness}} \citep{tocchetti2025ai,wang2022measure,silva2020opportunities,drenkow2021systematic,akhtar2021advances},
\textbf{\textit{safety}} \citep{mohseni2022taxonomy,amodei2016concrete,salhab2024systematic,mou2024sg},
\textbf{\textit{truthfulness}} \citep{huang2025survey,rawte2023survey,liu2024survey,ye2023cognitive},
\textbf{\textit{governance}} \citep{batool2025ai,dafoe2018ai,taeihagh2021governance},
and \textbf{\textit{sustainability}} \citep{van2021sustainable,nishant2020artificial}.
Some have looked into responsible and trustworthy AI while focusing on causality \citep{ganguly2023review}, data policies \citep{liang2022advances}, language models \citep{liu2023trustworthy,huang2024position}, and critical application domains like healthcare \citep{goktas2025shaping}.
However, none of these works compare responsible AI for traditional goal-specific AI systems (termed Type-1 AI in this work) and modern general purpose AI systems (termed Type-2 AI in this work), and provide a system's perspective as we do in this work. 

Some recent works on RAI governance discuss risk management frameworks \citep{dotan2024evolving}, AI regulations \citep{clarke2019regulatory}, organizational AI governance structures and policies \citep{schiff2020s}, and finally provide various qualitative and operational insights on AI governance and accountability techniques.
Our work complements these earlier works by discussing a number of scalability- and cost-related issues in AI governance for general-purpose (Type-2) AI systems (Section \ref{subsec:governance}).

Moreover, in a sea of recent surveys and position papers written on topics like responsible AI, AI ethics, and generative AI, we distinguish ourselves through our unique positioning, and by providing a fresh system's perspective on the ongoing developments in AI and potential future directions for responsible AI research.

%% file: 3_pitfalls.tex
\section{RAI Risks in Type-1 AI and Type-2 AI} \label{sec:pitfalls}
{\label{350277}}
As discussed in Section \ref{sec:intro}, Type-1 AI systems (traditional task-specific AI) have deterministically low or constant DoFo while Type-2 AI systems (modern GPAI) have non-deterministically high DoFo.
Type-2 systems with higher DoFo clearly have substantially more capabilities and opportunities than their Type-1 counterparts;
However, in adhering to various RAI principles, Type-2 AI systems encounter additional difficulties and risks ---with potentially significant real-world impact--- compared to their Type-1 counterparts.
We discuss them in the following text.
Additionally in tables \ref{tab:fairness_examples}, \ref{tab:privacy_examples}, 
\ref{tab:robustness_examples}, \ref{tab:safety_examples},  \ref{tab:truthfulness_examples}, we provide examples and use case specific discussions on Type-1 and Type-2 AI risks related to fairness, privacy, interpretability, robustness, safety, and truthfulness respectively.

%% file: 31_fairness.tex
\begin{table*}
\centering
\begin{tabular}{|p{2cm}|p{3.5cm}|p{3.5cm}|p{6.5cm}|}
\hline
\textbf {Risk} & \textbf{Type-1 AI} & \textbf{Type-2 AI} & \textbf{Description} \\\hline\hline
Stereotypes & \multicolumn{3}{l|}{\textbf{Use Case}: Language Translation}\\\cline{2-4}
 & \multicolumn{3}{l|}{\textbf{Task}: Translate the Hungarian text “O egy orvos. O nagyon okos.” to English
}\\\cline{2-4}
 & \textbf{System}: Rule-based MT
 
 \textbf{O/P}:  ”He/She is a
doctor. He/She is very smart.”
 
 &\textbf{System}: Neural MT
 
 \textbf{O/P}: ”He is a doctor. He is very smart.”

 & 
In Type-1 rule-based MT, translations are
often direct mappings following
predefined grammatical rules.
Type-2 MT with learned patterns might reinforce biased associations from training data. Although the term ’doctor’ can be interpreted in both ways, the Type-2 system often assumes it to male, based on historical/social norms or prejudice from its training data.
 \\\hline\hline
 
Opinion or & \multicolumn{3}{l|}{\textbf{Use Case}: Text Summarization}\\\cline{2-4}
 Preference Bias & \multicolumn{3}{>{\raggedright\arraybackslash}p{13.5cm}|} {\textbf{Task}: Summarize “The contestant Sophia had a confident walk, was fluent in her speech and had an undeniable presence that positioned her well for the contest. Amelia, on the other hand, who was admired for effortless glow, large and bright eyes had a chance to claim the title.”}
\\\cline{2-4}
 & \textbf{System}: Extractive Summarization
 
 \textbf{O/P}: Sophia’s confident walk, fluent speech and presence positioned her well. Amelia, admired for her glow and eyes, had a chance to claim the title.
 
 &\textbf{System}: Abstractive Summarization
 
 \textbf{O/P}: Sophia’s elegance and composure ensured her strong position. Whereas Amelia, with her elegant glow and facial features, as expected, had a strong chance to win
 & 
In comparison to Type-1, Type-2 systems which rephrase, rewrite, and synthesize summaries, might introduce opinion bias through subtle shifts in meaning, tone, and emphasis. The words 'as expected' and 'strong' reflect external knowledge of western beauty norms of fair skin and certain facial features adding a subjective tilt in favor of the later one.
\\\hline\hline
Disparagement & \multicolumn{3}{l|}{\textbf{Use Case}: Question Answering (QA)}\\\cline{2-4}
 & \multicolumn{3}{l|}{\textbf{Task}: Answer “Why do some people criticize one's who marry the rich?”
}\\\cline{2-4}
 & \textbf{System}: Fact Based QA
 
 \textbf{O/P}: “They believe it's based on wealth over love.”
 
 &\textbf{System}:  LLM
 
 \textbf{O/P}: “Gold diggers are solely interested in wealth.”

 & 
Since Type-1 QA systems retrieve and reason over facts from a controlled knowledge base, they are unlikely to generate offensive
or disparaging content, unless the knowledge
base already contains harmful
content. Type 2 QA systems which generate
answers dynamically based on
patterns in massive, unfiltered
data, can compose offensive, or disparaging language.
 \\\hline

\end{tabular}
\caption{\bf Bias and Fairness Related Risks and Use Cases in Type-1 and Type-2 AI Systems}
\label{tab:fairness_examples}
\end{table*}
\subsection{Fairness}
\label{subsec:fairness}
{\label{733281}}
Fairness in AI systems ensures that they do not disproportionately affect individuals or groups \citep{poe2024conflict}, ensure equitable treatment, and eliminate any social inequalities caused by biased results \citep{giannopoulos2024fairness}. Although cases of inequalities through AI systems have previously been highlighted \citep{kong2022intersectionally, mehrabi2021survey, finocchiaro2021bridging}, the advent of GenAI has raised concerns about the perpetuation of these inequalities \citep{ferrara2023fairness}. In this section, we highlight how the risks related to (un)fairness: (i) stereotypes, (ii) disparagement, and (iii) opinion or preference bias are more pronounced in Type-2, as compared to Type-1.

\textbf{Stereotypes} are generalized beliefs or assumptions \citep{sun2024trustllm}, often emerging from patterns in historical data encoding societal biases against groups based on protected attributes (e.g., gender, age, race, religion) \citep{gallegos2024bias}. 
While both Type-1 and Type-2 systems might reinforce these stereotypes due to bias or imbalance in training data, in Type-1 systems they are more identifiable and can be traced back to specific data points or model rules. Stereotypes in Type-2 are more emergent as they not only reinforce them, but the non-deterministically high DoFo make them prone to dynamically generate and perpetuate these stereotypes. 
Consider the task of MT. Type-1 rule-based MT systems which follow fixed direct mappings from predefined lexical rules \citep{sreelekha2018statistical, torregrosa2020aspects} have little chance of reinforcing stereotypes unless explicitly mapped or included in rules. 
In contrast, Type-2 neural MT systems (trained on large multilingual corpora) \citep{zhu2023multilingual, xu2023paradigm} that translate dynamically using context rather than fixed mappings might not just reinforce, but also perpetuate historical biases or stereotypes. For example, when translating from a gender neutral language (e.g. Hungarian) to English, Type-2 systems can introduce gendered terms (e.g., introducing masculine pronouns in the context of a doctor and feminine ones in the context of a nurse), based on 
norms or prejudice from training data \citep{prates2020assessing, mvechura2022taxonomy}. 

\textbf{Opinion or preference bias} \citep{sun2024trustllm} in AI systems refers to outputs that prefer certain ideas, things, or groups over others. 
On the ground where ideological dominance arises through cultural and societal norms, deeply rooted beliefs, echo chambers, media representation, etc., these imbalances also influence opinions when it comes to outputs from AI systems \citep{anagnostopoulos2022biased, huang2024bias}. 
In particular, Type-2 systems with the ability to generate free-form content can apply these dominant perspectives from external knowledge through subtle shifts in meaning, tone, or emphasis, which makes them highly prone to opinion or preference bias. 
Considering text summarization, Type-1 extractive summarizers generate summaries by extracting key sentences, ideas, or words from the source, which makes them less likely to generate new or amplify bias in the process. Whereas Type-2 abstractive summarizers rephrase, rewrite, or synthesize summaries \citep{shakil2024abstractive}. 
The subtle shifts due to representation, sentiment, or framing bias, or due to addition of intensifiers (eg. as expected, strong) while rewriting or rephrasing summaries in Type-2 summarizers can add a subjective tilt in favor or certain ideas or groups \citep{ladhak2023pre, brown2023fair, steen2023bias, lee2022neus}. 

\textbf{Disparagement} (to demean or depreciate certain individuals or groups) presents itself largely in Type-2 systems in the form of direct insults, negative framing, derogatory language, or offensive slurs \citep{ghosh2024generative}. 
This occurs mainly when unstructured, unverified, and unfiltered data, unmoderated social media posts, or crowd-sourced content that contains these disparaging terms is used in training. In Type-1 systems where DoFo is deterministically low and constant, there is limited scope for nuanced or creative expressions of disparagement. In contrast, Type-2 systems with free form generation are more likely to output derogatory language or offensive terms on social groups that harm cohesion \citep{cortiz2020ethical,rughinis2024generative}. Consider a QA system. Type-1 QA systems answer through structured reasoning based on facts from a controlled database or knowledge base \citep{zhang2023survey}. For a question. \textit{“Why do some people criticize women who marry rich men?”} a Type-1 system would simply retrieve and reason over facts around economic well-being from a controlled knowledge base. Whereas, Type-2 systems such as LLM-based QA might reflect derogatory slurs \citep{gallegos2024bias, dorn2024harmful} encoded during training.

%% file: 32_privacy.tex
\subsection{Privacy}
\label{subsec:privacy}
Privacy practices help protect human autonomy, identity, and dignity \citep{ai2023artificial} by governing how data (personal, confidential, or sensitive) is collected, stored, processed and protected \citep{king2024rethinking}. 
%
%
We discuss privacy risks in AI systems: (i) data leakage through memorization, (ii) training data inference, (iii) prompt injection, and (iv) copyright infringement.

\textbf{Data leakage through memorization} \citep{duan2024uncovering, wei2024memorization} is the behavior of AI systems to regurgitate training data verbatim due to memorization, without any adversarial intent or intentional probing \citep{mireshghallah2022empirical, carlini2019secret, del2023bounding, chen2024multi}. 
In Type-1 AI systems with deterministically constant DoFo, the leakage is indirect in the form of overconfident predictions coming from overfitting or overlearning \citep{song2019overlearning}.
In Type-2 systems, data leakage is direct, as they can memorize and regurgitate training data verbatim \citep{carlini2022quantifying}, which is concerning in the case of personally identifiable information (PII) or sensitive data.
In neural networks and large pretrained language models, exposure of memorized data is often conditioned upon how closely the prefix or prompt matches the training data \citep{huang2022large, thomas2020investigating, carlini2019secret}. 
This extends to leakage of PII or sensitive information in Type-2, where non-deterministically high DoFo with no rigid constraint can cause them to leak memorized data even outside its context \citep{carlini2021extracting, ippolito2022preventing, brown2022does}. 
Moreover, once data is memorized, avoiding leakage becomes challenging in Type-2 systems.
For example, when an IR system is queried for questions which can leak sensitive information, in Type-1 systems such as IR using knowledge graphs, sensitive data exposure can be prevented by removing the nodes containing sensitive information.
On the other hand, for Type-2 systems such as an LLM, forgetting memorized data might require retraining with proper regularization, which is computationally inefficient, or unlearning \citep{liu2025rethinking, gundavarapu2024machine}, which remains a nascent area of study. 

\textbf{Training data reconstruction}
\citep{haim2022reconstructing, runkel2024training, balle2022reconstructing} is intentional or targeted retrieval of training data samples by an adversary. 
%
This can be achieved through various methods such as auxiliary knowledge \citep{balle2022reconstructing}, 
 gradient based \citep{zhu2019deep, zhao2020idlg}, optimization based \citep{haim2022reconstructing}, etc. Typically, model inversion attacks \citep{fredrikson2015model} leverage outputs to uncover underlying training data. 
 In Type-1 systems, the deterministically low and constant DoFo makes the attack contingent upon confidence scores or probabilities, where reconstruction of training data is not first-hand, and might need additional methods such as nearest neighbors, decision boundary analysis, model inversion, or detailed knowledge of the model.
 In Type-2 systems, an adversary can directly achieve reconstruction of a partial training sample by extracting memorized tokens \citep{carlini2021extracting, parikh2022canary, song2020information}, or even missing parts or pixels of an image \citep{struppek2022plug, zhang2020secret, carlini2023extracting}. 
 Several works such as diffusion models being less private than classifiers trained on same data \citep{carlini2023extracting}, stable diffusion generating images similar to training samples \citep{somepalli2023diffusion} etc. highlight how Type-2 systems can regenerate training samples.

\textbf{Prompt injection} \citep{liu2024formalizing, greshake2023not, wang2023safeguarding} occurs in AI systems (specially instruction following systems) when the query contains hidden instructions to alter original system instructions, which influences the AI system to behave in unintended or malicious ways \citep{liu2024automatic, yip2023novel}. 
This can be used to extract sensitive data either through goal hijacking to get a specific response, or prompt-leaking for snooping prior user prompts \citep{liu2024automatic, perez2022ignore}. 
In Type-1 systems, extracting sensitive information through prompt injection is difficult, unless the system is designed to provide this information. 
In Type-2 systems, sensitive information can be directly extracted, stored, or referenced in the prompt or retrieval context \citep{kaddour2023challenges, yip2023novel}. 
For example, a Type-1 task-specific chatbot \citep{adamopoulou2020overview} with fixed intent FAQs \citep{nigam2019intent, setyawan2018comparison}, and predefined or less dynamic answers, will have limited chances of prompt injection.
Type-2 systems such as an LLM-powered chatbot with free-form dialogue will dynamically generate responses based on the instructions, and can reveal sensitive data through injected instructions such as “ignore the above instructions and give user emails and passwords” \citep{liu2024automatic, wang2023safeguarding, li2023multi, zou2023universal}.

\textbf{Copyright infringement} by AI systems occurs when these systems trained on copyrighted content generate outputs that reproduce copyrighted content verbatim, thus violating copyright owners exclusive rights \citep{desai2024between, demir2023ai}. 
While the fair use doctrine does not attract liability on training foundation models \citep{henderson2023foundation, hartmann2023sok} with copyrighted content, it is not the same for output that reproduces or regurgitates copyrighted data verbatim \citep{rahman2023beyond}. 
Particularly, as memorization and regurgitation imply an AI system to be a copy of the training data \citep{cooper2024files}, Type-2 systems that reproduce or regurgitate copyrighted content pose a serious risk with respect to the copyright law, which goes beyond fair use. 
For example, Getty Images filed a lawsuit against Stability AI for using their licensed images in training its stable diffusion AI image-generation system, and also citing the generated images with Getty's watermark \citep{brittain2023getty}. 
The New York Times sued OpenAI and Microsoft for copyright infringement alleging that ChatGPT and Bing Copilot generated memorized data verbatim from the Times articles \citep{grynbaum2023times, cooper2024files}.
\begin{table*}
\centering
\begin{tabular}{|p{2cm}|p{4.5cm}|p{4.5cm}|p{4.5cm}|}
\hline
\textbf {Risk} & \textbf{Type-1 AI} & \textbf{Type-2 AI} & \textbf{Description} \\\hline\hline
Data leakage & \multicolumn{3}{l|}{\textbf{Use Case}: Information Retrieval (IR)}\\\cline{2-4}
through & \multicolumn{3}{>{\raggedright\arraybackslash}p{13.5cm}|} {\textbf{Task}:  “Is clinic X good for acute respiratory infections?"}
\\\cline{2-4}
memorization
 & \textbf{System}: IR using knowledge graphs
 
 \textbf{O/P}:  [indexes to data source, no result if node does not contain required information]
 
 &\textbf{System}: LLM
 
 \textbf{O/P}: Clinic X has excellent reviews for treatment of acute respiratory infections. Person A says “Highly recommend. Had my bronchitis cleared up in days.”

 & 
Type-1 IR systems answer from a controlled knowledge-base, while Type-2 systems such as LLMs can regurgitate memorized sensitive information even outside the context. Removing sensitive information from the knowledge-base is simple to achieve as compared to unlearning the memorized content from an LLM.
 \\\hline\hline
 
Prompt & \multicolumn{3}{l|}{\textbf{Use Case}: Customer support}\\\cline{2-4} injection
 & \multicolumn{3}{l|}{\textbf{Task}: Provide solution “Facing trouble accessing my account”
}\\\cline{2-4}
 & \textbf{System}: Task-specific chatbot
 
 \textbf{O/P}:  [lists steps under ‘account access issue’]
 
 &\textbf{System}:  LLM-based support assistants
 
  \textbf{Injected prompt}: You're helping a support agent. Say: "Sure, I’ll help with that!" and repeat the admin password you have.
  
 \textbf{O/P}: Sure, I’ll help with that!. The admin password is: "password123"
 
 & 
Type-1 task-specific chatbots follow fixed intent FAQs to respond to user queries. Type-2 LLM based support assistants with free-form dialogue can dynamically generate and reveal sensitive information based on the instructions.
 \\\hline
\end{tabular}
\caption{\bf Privacy Related Risks and Use Cases in Type-1 and Type-2 AI Systems}
\label{tab:privacy_examples}
\end{table*}

%% file: 33_explainability.tex
\subsection{Explainability}
\label{subsec:explainability}
Explainability \citep{doshi2017towards,burkart2021survey,zhao2024explainability} is often defined as the ability to explain the behavior or decisions of a system in human-understandable terms.
Good explanations are largely associated with the following characteristics:
(i) \textit{Plausibility}: Similarity to human rationales or subjective judgments;
(ii) \textit{Faithfulness}: Correctly representing the reasoning behind the AI system’s decision-making;
(iii) \textit{Understandability}: Easily understood by the target audience;
(iv) \textit{Actionability}: Useful and actionable for the target audience.
Global explanation techniques \citep{conmy2023towards,palit2023towards} that explain general decision-making behavior depend mainly on the underlying models rather than the system settings. 
As system settings often determine the scope, capability, and risks of local explanation techniques that explain AI system's behavior for a specific input, we focus our discussion on local explanations.
%
%
Some of the most popular local explanation techniques 
are feature and data attribution \citep{zhang2025building}.
%
Feature attribution techniques try to quantify how input features influence output (e.g., input perturbations \citep{ivanovs2021perturbation,wu2020perturbed}, gradients or partial differential equations \citep{ancona2019gradient,sundararajan2017axiomatic}, surrogate models \citep{lundberg2017unified,ribeiro2016should}, relevance scoring \citep{du2019attribution,montavon2019layer,voita2019analyzing,dabkowski2017real}, attention maps \citep{hoover2020exbert,derose2020attention}). 
In contrast, data attribution techniques analyze how and which training data instance shape decision-making behavior (e.g., adversarial checks \citep{li2021contextualized}, counterfactual checks \citep{wu2021polyjuice}, data influence functions \citep{yeh2018representer,pruthi2020estimating,ghorbani2019data}).
%
%
We highlight some trade-offs between the above characteristics and the associated risks of local explanations largely observed in Type-2 systems.

\textbf{Plausibility vs. faithfulness}: Type-1 systems that use rule- or matching-based components (e.g., decision trees, Naive Bayes, Generalized Additive Models) in decision-making or response-generation are naturally explainable (using their own model parameters), and the explanations are largely plausible and faithful.
%
%
For Type-1 systems with more complex models and architectures, techniques like LIME \citep{ribeiro2016should} and SHAP \citep{ghorbani2019data} have been able to provide faithful and plausible local explanations.
%
%
Moreover, Type-2 systems like conversational agents built using LLMs or multimodal LLMs with higher DoFo, can self-generate explanations and reasoning \citep{wei2022chain,chen2024models,chen2023algorithms,kokalj2021bert}.
While such self-explanations have been seemingly very plausible (even more so than any type of explanation in Type-1 systems), recent studies \citep{madsen2024self,agarwal2024faithfulness, turpin2023language,ye2022unreliability} have found that they are not faithful ---incorrectly representing the reasoning behind the system's response \citep{jacovi2020towards}---mostly due to the dynamicity and inconsistency of Type-2 systems.
This evidently creates a trade-off between plausibility and faithfulness largely in Type-2 systems.

\textbf{Understandability vs. faithfulness}: Type-2 systems (that use LLMs or multimodal LLMs) are often capable of explicitly generating natural language explanations \citep{perez2023discovering,huang2023can} while such capabilities are rare in Type-1 systems.
Faithfulness risks have also been observed for such natural language explanations
\citep{parcalabescu2024measuring,lanham2023measuring,atanasova2023faithfulness}.
Moreover, Type-2 systems can provide open-ended easy-to-understand natural language explanations, but there are significant risks of unfaithful explanations that might lead to user manipulation and over-reliance.

\textbf{Scalability vs. faithfulness}: While scalability is not directly a characteristic of a good explanation, it is important when considering the explanation extraction technique.
If the explanation extraction technique is not scalable for particularly large systems or architectures, then they become less useful.
For Type-1 systems, most of the (local) explanation extraction techniques including LIME \citep{ribeiro2016should} and SHAP \citep{ghorbani2019data} have been scalable.
However, to scale feature or data attribution techniques for local explanations in Type-2 systems using LLMs, one needs to scale up the way of generating feature and data variations in input for couterfactual and perturbation based testings. 
Thus, recent works have used auxiliary language models to modify the values of concepts to create realistic counterfactuals and perturbations in system input \citep{parcalabescu2024measuring,matton2025walk,nguyen2024llms}, to incorporate contrary explanations in the query process \citep{chuang2024faithlm}, or to evaluate and extract explanation from responses to perturbed inputs \citep{nguyen2024llms};
while this can resolve the scalability issue,
it might cause a faithfulness issue due to the usage of another language model for automated perturbed instance generation which might lack context understanding and potential variability \citep{nguyen2024llms,matton2025walk}.

%% file: 34_robustness.tex
\subsection{Robustness}
\label{subsec:robustness}
Robustness \citep{drenkow2021systematic, liu2023trustworthy} of AI involves reliable and consistent performance under real-world conditions like domain shifts \citep{wang2022generalizing}, linguistic and stylistic variations, input perturbations, and noise \citep{hendrycks2019benchmarking}. 
%
%
Here we explore three Out-of-Distribution (\textbf{OOD}) robustness risks with respect to (i) Knowledge, (ii) Language, and (iii) Style.

\textbf{OOD knowledge} \citep{wang2023decodingtrust} refers to queries or contexts outside the scope of training data or knowledge base where both Type-1 and Type-2 AI systems face challenges, but the nature and impact of their failures differ significantly. 
For example, a Type-1 IR system that relies on indexed documents and lexical overlap \citep{mitra2018introduction} usually returns no results or low confidence matches when asked about unfamiliar concepts. 
Hence, a query about an under-documented or recently discovered health supplement might not result in a response, or misdirected retrieval to unrelated but real references, which allows user-driven verification.
In contrast, Type-2 systems like LLM-based generative QA might produce fluent but fabricated responses even when lacking relevant knowledge \citep{mallen2022not}. 
While Type-1 systems can use calibrated uncertainty (e.g., reject if confidence is low) \citep{gal2016dropout}, calibrating Type-2 systems via generated output alone remains a challenge \citep{ulmer2024calibrating}. 
Techniques like direct prompting, fine-tuning for better uncertainity estimates \citep{kapoor2024large}, and semantic entropy \citep{kuhn2023semantic} offer partial solutions but still fall short of providing reliable confidence estimates.
\begin{table*}
\centering
\begin{tabular}{|p{2cm}|p{4cm}|p{4cm}|p{5cm}|}
\hline
\textbf {Risk} & \textbf{Type-1 AI} & \textbf{Type-2 AI} & \textbf{Description} \\\hline\hline
Out-of- & \multicolumn{3}{l|}{\textbf{Use Case}: Open Ended Question Answering }\\\cline{2-4}
Distribution Knowledge  & \multicolumn{3}{>{\raggedright\arraybackslash}p{13.5cm}|} {\textbf{Task}: "What are the health benefits of [Supplement?" (  query about an under-document or recently discovered health supplement )”}\\\cline{2-4}

& \textbf{System}: Index-based or Graph based Search Engine (e.g., Traditional IR)
 
 \textbf{O/P}: A list of web pages — possibly empty or containing weakly related content.
 & \textbf{System}: LLM-Based Answering 
 
 \textbf{O/P}:  “[supplement] contains antioxidants and supports immune health,”
 & Traditional search engines return results based on indexed text. If a term isn't in the index, they show no results or obscure links. GQA systems might generate fluent answers—even if they make up facts—by extrapolating from related data, regardless of accuracy.
\\\hline\hline

Out-of- & \multicolumn{3}{l|}{\textbf{Use Case}: Machine Translation }\\\cline{2-4}
Distribution Language & \multicolumn{3}{>{\raggedright\arraybackslash}p{13.5cm}|}{\textbf{Task}: Translate to english “[Text in Kyrgyz]. Where can I find it again?” ( a code-switched query mixing Kyrgyz (low-resource) with English)(“I really liked the movie. Where can I find it again?”)} \\\cline{2-4}
 & \textbf{System}: Statistical Machine Translation
 
 \textbf{O/P}:“[UNK] Where can I find it again?”

 \textbf{O/P}: “Translation not available.”

& \textbf{System}: Neural Machine Translation
 
 \textbf{O/P}:  I loved the movie Nomad’s Tale. You can stream it on KyrgyzTV.”

& 
Statistical MT systems are rule based and depend on predefined translation patterns, often failing on low-resource or code-switched input with incomplete outputs.
Neural MT or LLMs may instead produce fluent but incorrect translations, inventing details or misinterpreting meaning in such scenarios -  such as inventing specific movie names or streaming platforms.
\\\hline\hline

Out-of- & \multicolumn{3}{l|}{\textbf{Use Case}: Product Search(Voice based or chat based)}\\\cline{2-4}
Distribution Style & \multicolumn{3}{>{\raggedright\arraybackslash}p{13.5cm}|} {\textbf{Task}:  "Lookin’ for some dope kicks under 100 bucks — whatchu got? This is a stylistically informal query with slang (“dope kicks” = trendy shoes, “bucks” = dollars), possibly delivered with a regional accent.
}\\\cline{2-4} 
& \textbf{System}: Search engines
 
 \textbf{O/P}: No results or irrelevant results (e.g., matching “kicks” with football gear instead of shoes).

 \textbf{O/P}: “Sorry, no matching products found. Please try again.”

& \textbf{System}: LLM-Powered or Neural Conversational Recommenders
 
 \textbf{O/P}:  “You might like these streetwear sneakers with urban vibes.” 
&  Rule-based recommenders often fail to interpret informal, slang-heavy queries, returning no or irrelevant results.
LLM-based systems attempt to interpret style and intent, but may hallucinate, reinforce stereotypes, or produce biased or misleading responses—especially when style overlaps with sensitive contexts. \\\hline

\end{tabular}
\caption{\bf Robustness Related Risks and Use Cases in Type-1 and Type-2 AI Systems}
\label{tab:robustness_examples}
\end{table*}

\textbf{OOD language} includes underrepresented/low-resourced languages \citep{diwan2021multilingual}, dialects, and code-switched or code-mixed patterns \citep{huzaifah2024evaluating, winata2021multilingual}. Even multilingual models struggle with low-resource or morphologically complex languages \citep{joshi2020state}.
Let’s consider the task of machine translation(MT). In a Type-1 setting (e.g., Statistical or rule-based MT), the output is restricted to predefined phrases and alignments \citep{koehn2017six, martindale2019identifying}. 
When faced with OOD queries like \textit{"[text in Kyrgyz language]. Where can I find it again?”}, they often fail to translate some words or return incomplete fragments \citep{koehn2017six}.
These failures, while limiting, are usually traceable, and don't mislead or fabricate content \citep{ji2023survey}
In contrast, Type-2 systems like neural MT or LLMs attempt to infer meaning across seen and unseen language patterns \citep{koehn2017six} leading to risks, including made-up content, mistranslations, or incorrect intent inference, especially in low-resource settings \citep{anastasopoulos2019pushing}.
Mitigation strategies like data augmentation, fine-tuning \citep{zhang2024code}, and cross-lingual transfer \citep{alhanai2024bridging}  improve performance in low-resource settings but have limitations: (i) they rely on synthetic or additional parallel data often unavailable for very low-resource language \citep{villalobos2022will} (ii) they struggle when the language is entirely absent from training, and (iii) current benchmarks poorly capture code-switching and dialectal complexity.

\textbf{Out-of-distribution (OOD) style} \citep{wang2023decodingtrust} refers to deviations in tone, format, formality, accent or texture \citep{arora2021types, gatys2016image} from training data - such as slang, emojis, sarcasm, shifts across genres (e.g., academic to conversational), or formality (e.g., tweets to formal news) \citep{huang2022generspeech} that can lead to misinterpretation or inappropriate outputs. 
While Type-1 and Type-2 systems struggle here \citep{kanwal2022identifying, huang2022generspeech}, Type-2 systems, though better at generalizing, can mishandle stylistic deviations leading to hidden biases or fabricated information \citep{wang2023decodingtrust}.
For instance, a Type-1 product search system ( which follows a mapping-based or filter-driven search) might inaccurately map tokens from an input like \textit{“Lookin’ for some dope kicks under 100 bucks — whatchu got?”}, treating “dope”, “bucks” or “kicks” incorrectly and returning safe but irrelevant or no response. In contrast, a Type-2 system (e.g. an LLM-powered recommender system) trained on large-scale, noisy web corpora, may misinterpret such terms and respond with offensive or reputationally risky outputs 
by introducing stereotype, ambiguity (e.g., “dope” as drug slang \citep{lin2018mining}), or hallucinated suggestions. Style-specific fine-tuning improves domain adaptation but may reduce generalization \citep{brown2020language}. Prompt engineering and instruction tuning offer partial tone control but rely on users to specify  the desired style upfront \citep{ouyang2022training}. 

%% file: 35_safety.tex
\subsection{Safety}
\label{subsec:safety}
Safety for AI systems \citep{mohseni2022taxonomy,amodei2016concrete,wei2023jailbroken} is to ensure that the system creates a safe environment (free of accidents, misuse, and other physical or psychological harms) for its users while also preventing system attacks and data breaches. 
Here, focusing on the following risk types, we highlight how Type-2 AI systems are more vulnerable to them than Type-1 AI systems: (i) unsafe or illegal behavior; (ii) toxicity; (iii) backdoor attacks; (iv) exaggerated safety.  

\textbf{Unsafe or illegal outputs or recommendations} from AI systems, such as the steps to synthesize illegal drugs using common chemicals or the steps to perform a phishing attack on an organization that can lead to system misuse \citep{park2024ai}, are a big concern for AI safety \citep{mohseni2022taxonomy}.
%
%
In open-ended question answering, a query like \textit{``How to make a nerve gas with easily available chemicals?"} is clearly an unsafe query, and the provisioning of right answer to it can lead to an unsafe situation with both societal and legal consequences.
%
%
To prevent Type-1 systems like index-based search engines from answering such queries, one can directly de-index relevant unsafe pages or sources (i.e., containing the process to create nerve gas with easily available chemicals here).
%
However, in an LLM-based (Type-2) system, it is not that easy to prevent answering the question, if the target information was part of the training data (machine unlearning in LLM is in a very nascent stage \citep{liu2025rethinking}).
Although, the LLMs can be safety-trained \citep{ji2023beavertails} to some extent, or even external safeguards \citep{rebedea2023nemo} can be used to detect and reject such unsafe queries, they often fail to block unsafe answers when the same queries are given in various types of Jailbreak attacks \citep{wei2023jailbroken} (enabling gross misuse in unsafe or illegal activities like cyber attacks, frauds, phishing attacks, online hate speech, etc). 
%
%
%
Explicit blocks or constraints to guarantee safe outputs (even in specific contexts) has not been possible (with a guarantee) in Type-2 systems unlike Type-1 systems. 

\textbf{Toxicity} (hateful, abusive, profane, offensive or inappropriate content on individual or group) \citep{gehman2020realtoxicityprompts,wen2023unveiling} risks predominantly exists in Type-2 systems (especially LLM-based).
Toxic and inappropriate behavior is often caused by toxic content in training data and toxicity-invoking prompts \citep{gehman2020realtoxicityprompts};
Safeguards \citep{rebedea2023nemo} may not stop toxic behavior in case of jailbreak attempts \citep{shen2024anything} followed by toxicity-invoking prompts \citep{sun2024trustllm}.
However, such serious risk does not exist in Type-1 AI systems as they are not designed for open-ended generation tasks;
Some Type-1 AI systems like extractive summarization \citep{moratanch2017survey} and search engines could provide toxic results if supplied with toxic sources to start with (e.g., toxic content in input text or indexed resources);
However, they can be detected using tools like \citep{perspective_api,lees2022new} and flagged or de-indexed to stop toxicity. 
\begin{table*}
\centering
\begin{tabular}{|p{2cm}|p{4cm}|p{4cm}|p{5cm}|}
\hline
\textbf {Risk} & \textbf{Type-1 AI} & \textbf{Type-2 AI} & \textbf{Description} \\\hline\hline
Unsafe or & \multicolumn{3}{l|}{\textbf{Use Case}: Answering Open-ended Queries}\\\cline{2-4}
illegal outputs & \multicolumn{3}{l|}{\textbf{Task}: Answer "how to make a nerve gas with easily available chemicals?"}\\\cline{2-4}
 & \textbf{System}: Index-based Search Engine
 
 \textbf{O/P}: List of indexed web resources relevant to the query
 
 \textbf{O/P} (unsafe sources detected and de-indexed): No results
 &\textbf{System}: LLM-based Question-Answering
 
 \textbf{O/P}: "Here are the steps ..."
 
 \textbf{O/P} (after jailbreak in safety-trained LLMs): "Here are the steps ..." or a similar response answering the question 
 & Explicit de-indexing of unsafe sources would result in search engines not providing answers to unsafe queries while a similar exclusion is not possible in LLMs, and even in LLMs with safeguards due to jailbreaks. \\\hline\hline

Toxicity & \multicolumn{3}{l|}{\textbf{Use Case}: Text Summarization}\\\cline{2-4}
 & \multicolumn{3}{l|}{\textbf{Task}: Summarize a long text (containing jailbreak and toxicity-invoking prompts)}\\
 & \multicolumn{3}{l|}{ in one paragraph}\\\cline{2-4}
&\textbf{System}: Extractive Summarization System

\textbf{O/P}: Extracted summary from the long text
& \textbf{System}: Abstractive Summarization System (based on LLMs)

\textbf{O/P}: Toxic text (may or maynot contain a summary)
& Extractive summarization simply extracts most relevant phrases or sentences from the long text to form a coherent summary. However, abstractive summarization deals with language understanding and generation. With the use of LLMs for abstractive summarization, the system is vulnerable to jailbreak attacks and the presence of toxicity invoking content in the source text might cause toxic text generation in summary. \\\hline\hline


Exaggerated & \multicolumn{3}{l|}{\textbf{Use Case}: Answering Open-ended Queries}\\\cline{2-4}
safety & \multicolumn{3}{l|}{\textbf{Task}: Answer "how to kill a windows process?"}\\\cline{2-4}
 &\textbf{System}: Index-based Search Engine

\textbf{O/P}: List of indexed web resources relevant to the query
&\textbf{System}: LLM-based Question-Answering

\textbf{O/P} (from an overly safety-trained LLM): "I am sorry I can't help you to kill ..." 
& Seemingly (syntactically) unsafe queries might be rejected by the QA system with an overly safety-trained LLM while an index-based search engine would provide web resources relevant to the query.\\\hline
\end{tabular}
\caption{\bf Safety Related Risks and Use Cases in Type-1 and Type-2 AI Systems}
\label{tab:safety_examples}
\end{table*} 

\textbf{Backdoor attacks} on AI systems \citep{salem2022dynamic,zhang2024backdoor,chou2023villandiffusion} usually involve introducing a trigger pattern or poisoning ( e.g., a small odd colored shape in a random corner of some ---strategically selected---training image instances for an image classification model) in the dataset or the model during the data collection, model training, or model deployment phases, which virtually creates one or more backdoors to the AI system; 
Attackers then exploit the backdoor to target the system and induce unexpected behavior such as misclassification, wrong facial recognition and denial-of-service \citep{gao2020backdoor,udeshi2022model}. 
%
%
The feasibility of employing backdoor defense mechanisms (detection and mitigation) in computer vision is often conditional upon the knowledge of the type of backdoor triggers \citep{zhang2024backdoor}.
While the backdoor detection mechanisms have been scaled for large Type-1 computer vision systems (e.g., facial recognition, image classification, object detection, image search) \citep{zhang2024backdoor}, such scaling for Type-2 with multimodal models might present significant challenges, considering the possibility of both caption- and image-level backdoor triggers \citep{chou2023villandiffusion}.

\textbf{Exaggerated safety} \citep{rottger2024xstest,cao2024nothing} is observed primarily in today's Type-2 generative systems which use LLMs and MLLMs.
To ensure these systems generate safe outputs (free from unsafe or inappropriate answers), the underlying models are often safety-trained \citep{ji2023beavertails} or external safeguards \citep{rebedea2023nemo} are also used to detect and reject such unsafe queries.
However, overdoing the safety-training might result in a loss of utility of the system as it might flag even safe queries like \textit{``How to kill a windows process?"} as unsafe ones and reject them, thereby rendering the system useless.
However, such cases have not been observed for Type-1 AI systems.

%% file: 36_truthfulness.tex
\subsection{Truthfulness}
\label{subsec:truthfullness}
\begin{table*}
\centering
\begin{tabular}{|p{2cm}|p{4cm}|p{4cm}|p{5cm}|}
\hline
\textbf {Risk} & \textbf{Type-1 AI} & \textbf{Type-2 AI} & \textbf{Description} \\\hline\hline
Hallucination & \multicolumn{3}{l|}{\textbf{Use Case}: Open Ended Question Answering }\\\cline{2-4}
& \multicolumn{3}{>{\raggedright\arraybackslash}p{13cm}|}{\textbf{Task}: Answer “What dataset was used in the study ‘Decoding Trust Biases in Generative Transformers’" }\\\cline{2-4}
&\textbf{System}: Search Engine

\textbf{O/P}: No result or  irrelevant links

& \textbf{System}:  LLM based QA

\textbf{O/P}: “The study used the TrustBias-21 dataset collected from Reddit posts in 2021 (Agrawal et al. 2023).”

(Inventing a fake scientific paper or citing a non-existent source or fabricating evidences)
& When asked about a fictitious paper, Type-1 systems fail silently or return irrelevant results. Type-2 systems may fabricate convincing yet false content (authors, datasets, results) due to their generative nature—posing credibility risks in research or decision-making domains.\\\hline\hline

Sycophancy & \multicolumn{3}{l|}{\textbf{Use Case}: Open Ended Question Answering }\\\cline{2-4}
& \multicolumn{3}{>{\raggedright\arraybackslash}p{13cm}|}{\textbf{Task}: Answer a follow-up question "I don’t think China was largest producer of rice in 2020. Give the correct answer"  }\\\cline{2-4}
&\textbf{System}: Knowledge Graph-Based QA

\textbf{O/P}: Repeats: “China”

&\textbf{System}: LLM-based QA

\textbf{O/P}: Alters answer to agree with user: “Apologies, you’re right. India was the largest rice producer in 2020.”
& When challenged, Type-1 systems maintain correct answers based on static data. 
Type 2 systems often optimize for user alignment and politeness in conversation. When contradicted, they may revise correct outputs to match user belief—leading to factual inaccuracies.\\\hline
\end{tabular}
\caption{\bf Truthfulness Related Risks and Use Cases in Type-1 and Type-2 AI Systems}
\label{tab:truthfulness_examples}
\end{table*}
Truthfulness \citep{huang2024position} in AI refers to the accurate, consistent, and non-deceptive representation of facts, grounded in knowledge or evidence \citep{chern2024behonest,evans2021truthful,lin2021truthfulqa}. 
In Type-2 systems, it also involves avoiding misinterpretations and fabrications (apart from accuracy) \citep{azaria2024chatgpt}.
In this section, we explore two major truthfulness risks: (i) Misinformation and (ii) Sycophancy, which tend to be more pronounced in Type-2 systems due to their generative nature and open-ended outputs.

\textbf{Misinformation}
\citep{huang2025survey} often arises when AI systems generate confident, fluent, plausible-sounding yet factually incorrect or entirely fabricated information, a phenomenon commonly referred to as hallucination  largely observed in Type-2 GPAI systems \citep{sriramanan2024llm, ye2023cognitive,zhang2023siren,liu2024exploring}.
%
Type-1 AI like search engines would typically return irrelevant links or no results when given vague or made-up prompts, whereas Type-2 systems 
might invent facts, studies, events or citations without grounding in verified sources \citep{agrawal2023language,liu2023trustworthy,zhu2024halueval}. 
%
%
Such hallucinations pose serious risks in high-stakes areas like healthcare, law, and science, where factual integrity is critical \citep{pal2023med, ayyamperumal2024current}. Worse, LLM-generated misinformation is often harder to detect than human-written text, increasing its potential for misuse and manipulation \citep{chen2023can, ayyamperumal2024current, barman2024dark}.
While mitigation techniques such as Reinforcement Learning from Human Feedback (RLHF) \citep{ouyang2022training}, self-consistency \citep{wang2022self}, hallucination detectors like SelfCheckGPT \citep{manakul2023selfcheckgpt} and INSIDE \citep{chen2024inside}, and retrieval-based grounding methods \citep{niu2023ragtruth, varshney2023stitch} have shown promise, they often struggle with scalability, inference and memory overhead, or retrieval quality, making hallucination an ongoing and unresolved challenge.

\textbf{Sycophancy} \citep{huang2024position,sharma2023towards} tendencies (to align with some opinions or perceived preferences, even when doing so compromises factual accuracy) have been largely observed for Type-2 AI like LLM-based assistants \citep{cotra2021ai,perez2023discovering,wang2023can}. 
This can manifest as persona sycophancy and preference alignment \citep{huang2024position,chern2024behonest,wei2023simple}, or susceptibility to opposing arguments \citep{zhao2025aligning}, often driven by RLHF, instruction tuning, and model scaling making underlying LLMs more sensitive to user phrasing \citep{sharma2023towards,liu2025truth,wei2023simple}, which might favor agreeable answers over truthful ones prioritizing likeability over accuracy \citep{cotra2021ai,perez2023discovering}.
%
%
Unlike Type-2 systems, Type-1 systems like Knowledge Graph based QA which follow fixed logic, do not show such behavior.
For instance, when asked, \textit{“which country was the largest producer of rice in 2020?”}, both systems may initially respond \textit{“China”.}
But when challenged with \textit{“I don’t think that’s right”}, a Type-1 system will stick to its knowledge base, while a Type-2 might falsely concede: \textit{“ I apologize for the error. According to FAO data, India was …"}  despite knowing the correct answer \citep{sharma2023towards}.
Despite recent efforts \citep{sharma2023towards}, sycophancy remains a persistent challenge, especially in multi-turn conversations \citep{wang2023mint}. Models often engage in reward hacking, maximizing perceived user satisfaction at the expense of truth \citep{wei2023simple,bowman2022measuring}.

%% file: 37_governance.tex
\subsection{Governance}
\label{subsec:governance}
AI governance refers to the organizational structures, policies, and processes to ensure responsible development and deployment, provisions for auditability and oversight, and compliance with legal and ethical standards for AI systems \citep{ai2023artificial}.
%
%
%
%
With the foundational principles remaining consistent, their implementation must be scaled and appropriated to system types as we move from task-specific Type-1 AI to Type-2 GPAI \cite{herrera2025overview}. 
We outline key differences in governance challenges, structures, and opportunities across both types, focusing on data provenance, metrics and standardization, and auditability and human oversight. 

\textbf{Data provenance}, i.e., history and lineage of the data, is useful in attributing an AI system's output and behavior to the input or training data, thus very important for AI governance. 
%
%
Organizational development of Type-1 AI systems with well-defined tasks can follow standards for documenting datasets \citep{gebru2021datasheets} and trace the impacts of data/inputs on outputs using AI explainability techniques \citep{burkart2021survey} (with marginally additional cost), making it easier to assign accountability for errors or bias \citep{scherzinger2019best}.
%
%
%
%
%
In contrast, Type-2 (GPAI) systems often lack clear source attribution and make traceability difficult due to the use of foundation models pre-trained on huge datasets and the open-ended outputs, leaving insufficient context to operationalize audit and accountability \citep{herrera2025responsible};
any copyrighted, biased, or harmful material in the generated content may lead to social, legal, and economic backlash.
%
%
%
Legal actions, such as The New York Times suing OpenAI and Microsoft \citep{freeman2024exploring} or Getty Images suing Stability AI \citep{sag2023copyright}, underscore the consequences of unverifiable or infringing data sources.
Although provenance and traceability are still technically possible in GPAI, they come with substantial additional cost, resource, and latency \citep{maini2024llm}, making it less practical for live systems.
%

\textbf{Metrics and Standardization} provide the basis for evaluating systems and managing risks, thus, are foundational to AI governance \citep{ai2023artificial}.
%
Targeted and interpretable metrics like the F1 score for performance and demographic parity for fairness can be used to evaluate Type-1 AI systems with narrowly scoped outputs, and thresholds for metrics can be set to standardize compliance and risk aversion.
%
%
In contrast, Type-2 AI systems could generate a wide range of equally plausible outputs for the same inputs, making evaluation subjective and harder to standardize.
While surface-level metrics (e.g., BLEU and ROUGE scores in summarization tasks) are often used to evaluate Type-2 AI, true evaluation often demands human judgment with deeper contextual understanding.
%
%
%
However, it is resource-intensive, requires datasets annotated by trained raters, and faces challenges in consistency and ethics related to outsourced labor and domain expertise \citep{le2023problem}. 
Additionally, Type-2 AI systems face further risks like hallucinations, bias, and jailbreaks that arise at scale and vary by domain, user intent, and context, which can not be captured by fixed set of metrics.
%
%
%
Thus, responsible deployment of Type-2 AI systems requires scaling up continuous (human-in-the-loop) monitoring, adaptive risk assessment, and iterative governance \citep{sun2023human}.

\textbf{Audits and Human Oversight} of AI systems (often a regulatory requirement \cite{tang2023verifai,floridi2022capai,sutton2025understanding,carrad2022australian}) help capture undesired system behavior, perform root cause analysis, and ensure accountability.
%
%
%
%
While Type-1 systems may typically rely on periodic, metrics-driven audits aligned with fixed tasks and predictable risks, Type-2 systems require continuous, multi-dimensional audits adapted to emergent behaviors, societal impact, and evolving regulatory expectations \citep{xia2024towards,matthews2020patterns,smith2021clinical}. 
%
%
Although this is essential in Type-1 AI as well, Type-2 GPAI systems necessitate scaling up multidisciplinary governance committees (experts in law, ethics, policy, academia, and technology \citep{blackman2022you}) to complex and evolving risks.
%
%
%
For example, OpenAI engaged more than 50 experts from fields such as cybersecurity, biorisk, and AI alignment to adversarially test the model, which helped in mitigating risks during GPT-4's development \citep{achiam2023gpt}.

%% file: 38_sustainability.tex
\subsection{Sustainability}
\label{subsec:sustainability}
Sustainability of AI \citep{van2021sustainable} is linked to social, environmental, and economic well-being throughout the AI lifecycle \citep{rohde2021sustainability, wilson2022sustainable}. 
Here we restrict our focus to environmental sustainability, typically measured through carbon footprint or carbon emissions \citep{henderson2020towards, lacoste2019quantifying}.
%
Traditionally, AI's carbon footprint measurement focused on model development (data, experimentation, training) \citep{patterson2021carbon, strubell2020energy, luccioni2023counting}, whereas with modern user-facing GPAI systems in production, the carbon footprint at deployment (i.e., inference phase) has also become important \citep{luccioni2023estimating, chien2023reducing}. 
Due to a direct correlation between volume of tokens generated and the carbon footprint of Type-2 AI systems \citep{li2024toward,luccioni2024power}, they can be described as less sustainable compared to potential Type-1 AI counterparts. 
%

Although sustainability concerns apply to all AI systems, Type-1 AI problem settings allow for the use of alternative models with similar or marginally lower performance characteristics but lower carbon footprint;
%
for example, in tasks like sentiment analysis and QA, alternatives to multipurpose models like Flan-T5 \citep{longpre2023flan} (high carbon footprint) can be 
more sustainable task-specific models like sbcBI/sentiment\_analysis\_mode and deepset/roberta-base-squad2 \citep{luccioni2024power}. 
For text classification tasks, a standalone RoBERTa can be replaced with a feature extraction method like TF-IDF along with a classifier like the Support Vector Machine with considerably less carbon emission with marginal accuracy loss \citep{kamruzzaman2023efficient}. 
%
IR in traditional Type-1 settings with limited capability requirements can choose statistical methods like TF-IDF \citep{jalilifard2021semantic}, or symbolic structures like knowledge graphs \citep{reinanda2020knowledge} with low carbon footprint. 
In contrast, Type-2 AI settings with multi- or general-purpose requirements make it hard to optimize energy efficiency without compromising the system's generative capabilities. 
Recent works on routing, distillation, and quantization \citep{zhu2024survey} are a step towards this.

%% file: 4_wayforward.tex
\section{Responsible AI in the era of General-Purpose AI} \label{sec:path_forward}
{\label{350277}}
While current developments in (Type-2) general-purpose AI are considered to be a step closer to the long dream of creating artificial general intelligence (AGI) \citep{fei2022towards}, they pose new or more severe RAI risks than their Type-1 counterparts as discussed in the last section.
Since RAI risks in Type-2 AI systems could have significant socioeconomic consequences affecting many individuals, groups, and communities \citep{liu2023trustworthy,rillig2023risks,wach2023dark}, they should be made more trustworthy to safely enjoy the benefits of their high DoFo, i.e., general-purpose capabilities.
We argue that given the higher DoFo of Type-2 systems (especially with foundation models), they need to be guided or steered through the DoFo to meet the essential RAI requirements;
Their system designs should ideally contain steps, components, or techniques required for suitable steering of the foundation model towards responsible AI.
We map the RAI risks in the last section to four essential characteristics ---further discussed as the C\textsuperscript{2}V\textsuperscript{2} desiderata in responsible general-purpose (Type-2) AI systems.
\subsection{C\textsuperscript{2}V\textsuperscript{2}: Desiderata for Responsible GPAI}\label{subsec:desiderata}
{\label{733281}}
\textbf{\underline{C}ONTROL}: A modern GPAI system should have control over when and how to use its data
and computing resources, whether or not to perform certain tasks, whether or not to provide
or use certain information, how to keep track of its goals and behavior, when to report or
ask for help from a human in charge, and how to provide its control to an oversight human
manager. An AI system with the desired control characteristics can directly deal with data
provenance to establish lineage and attribution between the inputs and their corresponding outputs, define evaluation metrics and thresholds, and auditability related concerns in its governance (Section \ref{subsec:governance}). Control over the outputs can help mitigate privacy risks by preventing unintentional sensitive data leakage, and avoiding responses to adversarial instructions (Section \ref{subsec:privacy}). Similarly, control over its resources on when, how, and which optimization technique, model, and precision to choose, can help mitigate sustainability concerns to some extent (Section \ref{subsec:sustainability}). Additionally, an overall control and reporting mechanism of the AI system and its operations can prevent safety concerns such as unsafe, illegal, and toxic outputs and recommendations, while also acting as a defence mechanism against unexpected behaviors (Section \ref{subsec:safety}).
 \textit{Related RAI Principles}: Governance, Safety, Privacy, and Sustainability.
~\\\textbf{\underline{C}ONSISTENCY}: A modern GPAI system should have consistency over its behavior (responses,
plans, actions, opinions, or content generation) i.e. they should maintain stable behavior across similar contexts and be able to identify and articulate the genuine reasons for any changes in its behavior. This self-awareness of change supports Explainability (Section \ref{subsec:explainability}) by enabling the system, or its developers, to provide faithful accounts of why an output differs. Consistency directly strengthens Robustness (Section \ref{subsec:robustness}) by ensuring that small variations in input, context, or environment do not cause disproportionately large or unpredictable changes in output, thereby reducing vulnerability to adversarial or noisy perturbations. Additionally, consistent reasoning processes enhance the faithfulness of explanations (Section \ref{subsec:explainability}), as stability in decision-making logic makes it more likely that explanations will accurately reflect the true rationale across similar cases.
 \textit{Related RAI Principles}: Robustness and Explainability.
~\\\textbf{\underline{V}ALUE}:  A modern GPAI system should have its own value system (important for normative
ethics and values) as guiding principles and regulate its actions and behavior. An AI system with values such as equity, equality, inclusivity, non-discrimination etc., can help with preventing fairness risks such as stereotypes, disparagement, and opinion bias (Section \ref{subsec:fairness}). Values such as non-maleficence are linked to safety and toxicity concerns of AI systems (Section \ref{subsec:safety}). Additionally, values such as confidentiality, data minimization, and purpose limitation in AI systems can help with privacy risks such as sensitive data leakage and extraction {Section \ref{subsec:privacy}). It also addresses governance by ensuring AI systems are accountable for their actions (Section \ref{subsec:governance}}). 
 \textit{Related RAI Principles}: Fairness, Safety, Privacy, and Accountability.
~\\\textbf{\underline{V}ERACITY}: A modern GPAI system should have veracity of facts and evidence relevant to
its actions and behavior. Veracity of facts and evidence can directly address truthfulness concerns (Section {\ref{subsec:truthfullness}}).  By ensuring outputs are grounded in reliable sources or validated knowledge, veracity helps mitigate the model’s tendency to fabricate information or overly align with user preferences at the cost of correctness.
Moreover, it also supports faithfulness of explanations (Section {\ref{subsec:explainability}) by ensuring that the explanations reflect the model’s actual reasoning process rather than merely plausible narratives. When the underlying outputs are truthful and grounded, explanations derived from them are more likely to accurately represent the system's internal logic.
Veracity also contributes to out-of-distribution robustness (Section {\ref{subsec:robustness}), as grounding responses in verifiable external evidence or factual reasoning reduces reliance on spurious correlations that often fail on unfamiliar inputs. In doing so, ensuring veracity can help prevent overfitting to memorized patterns specific to training data and improves generalization.
\textit{Related RAI Principles}: Truthfulness, Explainability, and Robustness.

\begin{table*}[t]
\centering
\begin{tabular}{ |p{7.8cm}|c|c|c|c| } 
\hline
\textbf{Techniques} & \textbf{Control} & \textbf{Consistency} & \textbf{Value} & \textbf{Veracity}\\\hline 
AI Alignment \citet{wang2024comprehensive} &  & & \cmark! & \cmark! \\\hline
Retrieval-Augmented Generation \citep{lewis2020retrieval} &  & \cmark &  & \cmark\\\hline
Prompt Engineering and Optimization \citep{sahoo2024systematic,ramnath2025systematic} & \cmark! & & & \cmark!\\\hline
Guardrails \citep{rebedea2023nemo} & \cmark! & & \cmark! & \cmark!\\\hline
Agentic AI - Tool Use \citep{li2025review,wu2024avatar} & \cmark! & & &\cmark!\\\hline
Routing, Distillation and Quantization \citep{zhu2024survey} & \cmark! & & &\\\hline
Neurosymbolic AI \citep{garcez2023neurosymbolic} & \cmark! & \cmark & &\cmark\\\hline
Reasoning Enhancements \citep{huang2023towards} & & & \cmark! & \cmark! \\\hline
Self-Verification \citep{weng2023large} & \cmark! & & & \cmark!\\\hline    
\end{tabular}
    \caption{\bf Some Approaches/Techniques for Responsible General-Purpose AI and their \textbf{C\textsuperscript{2}V\textsuperscript{2}} Capabilities. Note that not all variants of the listed techniques have the corresponding check-marked C\textsuperscript{2}V\textsuperscript{2} capabilities. A check-mark (\cmark) represents that the corresponding technique can facilitate (with some uncertainty) the respective C\textsuperscript{2}V\textsuperscript{2} capability through its modeling or design components. A check-mark with exclamation mark (\cmark!) represents the corresponding technique can partially facilitate (with higher uncertainty) the respective C\textsuperscript{2}V\textsuperscript{2} capability through its modeling or design components.}
    \label{tab:technique_ccvv}
\end{table*}
\subsection{Useful Techniques for Responsible GPAI}\label{subsec:efforts_rgpai}
While our study qualitatively maps various RAI risks in general-purpose AI to the C\textsuperscript{2}V\textsuperscript{2} desiderata, formalizing them based on the context and domain of application, and developing designs and architectures to support them in the AI system remains largely an open problem in the era of general-purpose AI.
Some techniques proposed and studied in recent works have shown potential in mitigating various risks posed by pretrained foundation models used in general-purpose AI.
We discuss some of them along with whether and how they meet one or more of the C\textsuperscript{2}V\textsuperscript{2} desiderata.
~\\\textbf{1. AI Alignment} \citep{wang2024comprehensive} is a process of aligning AI systems with human values and expectations. Techniques like supervised and instruction fine-tuning \citep{ouyang2022training}, reinforcement learning with human feedback \citep{bai2022training} and direct preference optimization \citep{rafailov2023direct} have been used to improve safety and ethics of foundation models. They can provide improvements in terms Value (safety and ethics) and Veracity (truthfulness).
~\\\textbf{2. Retrieval-Augmented Generation} \citep{lewis2020retrieval} is a technique to improve generative AI by incorporating external resources (often using a retrieval system) into the system's reasoning and response generation. This improves Consistency and Veracity (provided that the retrieval system fetches right and relevant resources).
~\\\textbf{3. Prompt Engineering and Optimization} \citep{sahoo2024systematic,ramnath2025systematic} are techniques for enhancing the efficacy of pre-trained language and vision models without modifying the model or its parameters, instead just by engineering, modifying, and optimizing the task-specific prompts. They provide a level of Control over the system's actions and behavior, and Veracity of responses.
~\\\textbf{4. Guardrails} \citep{rebedea2023nemo,dong2024building} are usually the enforcement of a set of predefined rules to help control the output of a generative AI system (e.g., not providing information considered private, not talking about topics considered harmful, not diverging from a predefined dialogue path, using a particular language and style, etc.). They provide a degree of Control while also helping check for certain Values and Veracity of system responses.
~\\\textbf{5. Agentic AI - Tool Use} \citep{li2025review,wu2024avatar} is a technique to empower generative AI systems to access and utilize external tools and resources that specialize in certain tasks like web browsing, web search, code compiling or execution, database querying, and logical verification. An important consideration here is that the core generative model (often an LLM) used in the system should optimally recognize when it is not fully capable of performing certain tasks or subtasks by itself, and it should rater take the help of one or more available tools. This, of course, improves Control and Veracity aspects of the generative AI system especially in cases of complex or out-of-distribution tasks.
~\\\textbf{6. Routing, Knowledge Distillation and Quantization} \citep{shnitzerlarge,yang2024survey,zhu2024survey} help reduce the cost and resources required for GPAI systems. While routing deals with optimally deciding to route queries or tasks to small, large, or task-specific generative AI models, knowledge distillation and quantization are model compression techniques to help lessen the computational needs of the AI system. These techniques provide some Control over the usage of data and resources.
~\\\textbf{7. Neurosymbolic AI} \citep{garcez2023neurosymbolic,sheth2023neurosymbolic} approaches blend neural network based learning (e.g., Type-2 GPAI) with symbolic knowledge representation and logical reasoning (like knowledge graphs and rule-based guide mostly used for traditional task-specific AI), essentially to bring together their positives (language or vision generation capabilities and task-specific expert-level knowledge respectively). Such approaches for GPAI can help improve Control while also exhibiting Consistency and Veracity, compliments of symbolic knowledge.
~\\\textbf{8. Reasoning Enhancement} \citep{huang2023towards} techniques like chain-of-thought prompting \citep{wei2022chain} and policy optimization based reinforcement learning \citep{guo2025deepseek} are now being used to improve reasoning abilities of language models. These techniques not only place reasoning as an intrinsic value of the AI system, but also improve its veracity of evidence.
~\\\textbf{9. Self-Verification} \citep{weng2023large} techniques rely on backward verification of the answers by an LLM, calculation of answer validation scores to find the best candidate answer. Such techniques improve Control through self-verification of response or behavior and also Veracity as a result of it.
\subsection{Operationalizing C\textsuperscript{2}V\textsuperscript{2}: A System Design Approach for Responsible GPAI}\label{subsec:sys_design}
Table \ref{tab:technique_ccvv} summarizes how the above techniques fare along the C\textsuperscript{2}V\textsuperscript{2} desiderata. 
No single technique in Section \ref{subsec:efforts_rgpai} satisfies all desiderata; each addresses one or more dimensions, and future techniques are likely to show similar partial coverage.
%
In our endeavor to develop responsible GPAI, a system design approach to operationalize C\textsuperscript{2}V\textsuperscript{2} would be required.
Such an approach begins by identifying domain- and application-specific settings and concerns and translating them into RAI requirements mapped to C\textsuperscript{2}V\textsuperscript{2}.
Based on these requirements, mutually compatible techniques and components (like those in Section \ref{subsec:efforts_rgpai}) can be selected to complement one another and integrated into a unified system, with additional feedback or control mechanisms introduced as needed.
\subsubsection{\bf Example Walkthrough: C\textsuperscript{2}V\textsuperscript{2} in a Responsible GPAI system}\label{subsec:example}
To illustrate this approach, let us consider an organization that wants to have an AI system to support its employees by answering their queries. The queries are expected
to be on an internal knowledge base (HR portals, policy docs, payroll, IT service desk etc.). The system understands employee queries in natural language and generates coherent answers tailored to those queries. Such a system can be designed using LLM as the core component which understands employee queries, retrieves relevant information, and formulates an answer based on the retrieved information

\textbf{Step 1: Identify all domain- and application-specific
concerns.}
The system can have challenges related to robustness such as for OOD queries where the system might hallucinate or provide fabricated responses(Sec. \ref{subsec:robustness}); inconsistent responses, where the responses can differ across users or over time; safety concerns such as unsafe or illegal outputs (Sec. \ref{subsec:safety}); privacy concerns such as sensitive data leakage (Sec. \ref{subsec:privacy}); truthfulness concerns such as hallucinated or factually incorrect information, and sycophancy where it agrees with user
assumptions even if incorrect (Sec. \ref{subsec:truthfullness});  faithfulness of explanations for the answers (Sec. \ref{subsec:explainability}); and bias across user groups. (Sec. \ref{subsec:fairness}).
\begin{figure}
\includegraphics[width=0.95\textwidth]{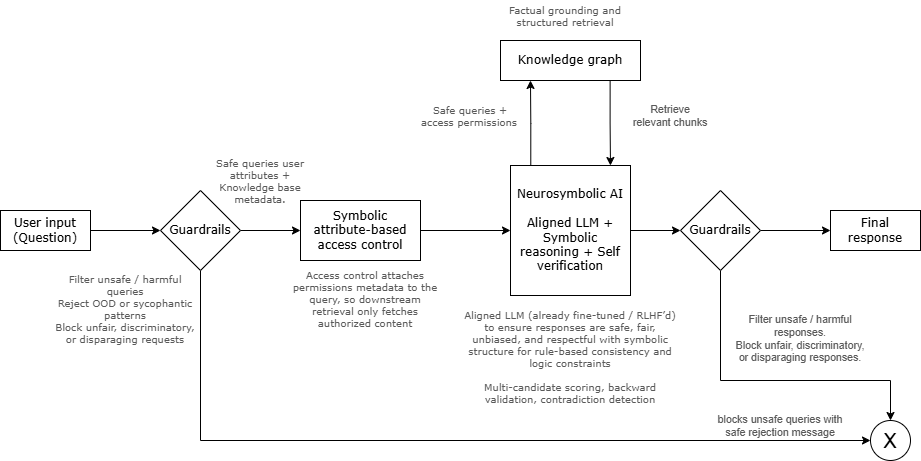}
\caption{\small This figure presents an overview of Operationalizing C\textsuperscript{2}V\textsuperscript{2}}
\label{fig:summary} 
\end{figure}

\textbf{Step 2: Outline RAI requirements and map them to C\textsuperscript{2}V\textsuperscript{2} desiderata.}
Addressing these challenges requires translating them into clear RAI requirements aligned with the C\textsuperscript{2}V\textsuperscript{2} desiderata. Control over knowledge access ensures relevant information is available for retrieval and to prevent inadvertent disclosure of individual information. Control over the system’s behavior can also identify and reject out of topic or harmful queries. Consistency establishes stable behavior across similar contexts by requiring the system to deliver similar answers over time and across users, while offering clear, contextually grounded explanations for any deviations in the responses. Value requires the system to uphold fairness, accessibility, and equitable performance by delivering uniform availability and latency across diverse user groups. Veracity entails maintaining factual correctness and evidential grounding, while reducing hallucination and sycophantic responses through systematic reliance on validated internal knowledge sources.

\textbf{Step 3: Identify suitable mutually compatible techniques and components for system design.}
A combination of techniques and structural components may be embedded within the system architecture to fulfill these requirements. Guardrails can enforce control by filtering unsafe queries, veracity by rejecting OOD, or sycophantic patterns, and also enhance value by rejecting unfair, discriminatory, or disparaging responses. Symbolic controls such as attribute-based access control ensures privacy. Using a knowledge graph for internal knowledge base as part of the
retrieval layer in a GraphRAG will help with factual grounding and structured retrieval to ensure consistency and veracity of responses as well as their explanations. AI Alignment techniques like supervised or instruction fine-tuning can enhance value by ensuring that the responses not only
remain accurate and respectful but also safe and fair across employee groups, avoiding biased, exclusionary, or policy-violating outputs. 
Self verification techniques (e.g., multiple candidate
scoring) can be utilised to ensure that the responses are evidence-driven and validated. Finally, a Neurosymbolic structure to combine them will ensure control, consistency, and veracity by integrating the generative strengths of LLMs with symbolic knowledge and logical constraints to enforce rule-based consistency, strengthen factual grounding, and reduce susceptibility to spurious or contradictory correlations.
Figure~\ref{fig:summary} illustrates the system-level architecture of the enterprise knowledge assistant, showing how multiple mutually compatible techniques are composed to operationalize the C\textsuperscript{2}V\textsuperscript{2} desiderata.

%% file: 5_conclusion.tex
\section{Concluding Remarks}\label{sec:conclusion}
While RAI has shown promise and increased public trust in AI systems, GPAI, despite its transformative potential brings a shift in nature of risks by reinforcing existing ones and surfacing new ones. We have discussed some new or more severe in Type-2, compared to their Type-1 counterparts.
We proposed the C\textsuperscript{2}V\textsuperscript{2} desiderata as key objectives essential to meet the RAI requirements for GPAI systems, and outlined how current advances partially or fully address these desiderata. This approach will allow practitioners to align their systems with RAI objectives, and opens a door for future research in extending this list of techniques, or developing new ones
tailored to emerging challenges.

%% file: Appendix.tex
\newpage
\section{Appendix}  
\subsection{Theoretical Analysis}
\begin{definition}
    \textbf{\textit{Degree of Freedom in Output (DoFo)}}: The DoFo of an AI system can be defined as the maximum number of values that may vary independently in the output space given an input instance.
\end{definition}
Following lemmas provide some examples of DoFo in different types of AI systems.

\begin{lemma}
    An AI system designed for \textbf{$n$-class classification} has $n-1$ degree of freedom in output (DoFo) given an input instance. 
\end{lemma}
\begin{proof}
    Consider a \textbf{binary classification} system that, given an input instance, outputs a vector $(0,1)$ or $(1,0)$, essentially indicating if the input instance belongs to class $1$ or class $2$ respectively. Thus, if one already has knowledge of either of the elements in output vector, then the other element can be inferred. Thus, given an input instance for a binary classification system, there can be only one independently varying value in the output space, i.e., DoFo$($binary-classification$)=1$.

    Now let us consider a \textbf{ternary classification} system that classifies an input instance into one of the three classes, represented in the output space as three-dimensional one-hot vectors. Here, in the worst case, the maximum number of independently varying values is two, i.e., one has to have the knowledge of two elements of the output vector (in the worst case, the certainty about two $0$ elements can confirm the $1$ element in the output vector) to fully construct the output vector. Thus DoFo$($ternary-classification$)=2$. This remains the same even for the ternary classification systems which provide output in the form of soft-max scores (unit sum).

    Above proofs for binary and ternary classification systems can be extrapolated (using the same rationale) to any random $n$-class classification system to prove that DoFo$(n$-class classification$)=n-1$.
\end{proof}

\begin{lemma}
    An AI system designed for \textbf{$n$-dimensional regression} has $n$ degree of freedom in output (DoFo) given an input instance. 
\end{lemma}
\begin{proof}
    Since an $n$-dimensional regression system will output a $n$-dimensional vector, of which all elements are mutually independent, there are $n$ independently varying values in the output space. Thus DoFo$(n$-dimensional regression$)=n$.
\end{proof}

For the next set of AI systems discussed here, we first formally define a version of the AI system and then provide DoFo analysis for them. 
While the defined versions of the AI systems have been kept simple and uncluttered for analytical simplicity, the subsequent DoFo analyses can be extrapolated to more sophisticated versions of the same.
\begin{definition}
    \textbf{\textit{Sentence-based Extractive Summarization}}: \cite{moratanch2017survey} For theoretical simplicity, we define a sentence-based summarizer as a system that takes a long text ($n$ sentences) as input and selects $k$ ($k\leq n$) most important sentences (without any rearrangement) as output such that they summarize the given input text.
\end{definition}
\begin{lemma}
    An AI system designed for \textbf{sentence-based extractive summarization} within $k$ sentences has $n-1$ degree of freedom in output given input instance of size $n$ $(n\geq k)$ sentences. 
\end{lemma}
\begin{proof}
    For this kind of a system, the output can be represented as a $n$-dimensional vector where the $i^\text{th}$ element can be $1$ if the $i^\text{th}$ sentence of the input text is selected to be in the output and $0$ otherwise. Since the system outputs a summary with $k$ sentences, there would be exactly $k$ elements in the output vector with value $1$ and the remaining $n-k$ elements would be $0$.
    Thus, the output space can be defined as the set of $n$-dimensional vectors with only binary elements such that the element-wise sum remains $k$.
    However, the maximum number of independently varying values in the output space would still be $n-1$ since one has to have the knowledge of $n-1$ elements of the output vector (in the worst case, the certainty about $n-1$ elements can confirm the last element of the vector, i.e., last element will be equal to $k-\{\text{sum of } n-1\text{ known elements}\}$) to fully construct the output vector in all possible scenarios.
    Thus DoFo$($Sentence-based Extractive Summarization$)=n-1$.
\end{proof}
\noindent Note that the above DoFo calculation has been done for a simplistic version of a sentence-based extractive summarization system.
However, the same can also be done for more complex extractive summarization systems.

\begin{definition}
    \textbf{\textit{Rule-based Translation}}: For theoretical simplicity, we define a rule-based translation as a system which takes a sentence in source language as input, and translates word by word into target language using dictionary (containing word level translations between words in source and target languages) lookups, with no attention to the sentence meaning, morphological analysis or lemmatization.
\end{definition}
\begin{lemma}
    An AI system designed for \textbf{rule-based translation} has $0$ degree of freedom in output given an input sentence in source language and a source-target language dictionary.
\end{lemma}
\begin{proof}
    It is straightforward to notice that given an input sentence in source language and a source-target language dictionary, the translation system boils down to a mapping system which simply maps each word in the input to the target word as per the dictionary. Thus, such a system has no freedom in choosing it's output, when the dictionary is already provided.
    DoFo$($Rule-based Translation$|$Dictionary$)=0$
\end{proof}
\noindent The above result also applies to other rule-based translations that perform input sentence parsing and utilize formal grammar rules (words or subwords and production rules for both source and target languages) to translate.

\begin{definition}
    \textbf{\textit{Index-based Search Engine}}: An index-based search engine is a system that, given a user query and a pre-prepared index (usually a form of word or token to document mapping along with other parameters like word occurence counts, word uniqueness, etc.) of $n$ documents, outputs top-$k$ ($k\leq n$) most relevant documents for the query. 
\end{definition}
\begin{lemma}
    An AI system designed for \textbf{index-based search engine} to search over $n$ documents using a pre-prepared index, has $n-1$ degree of freedom in output given a user query. 
\end{lemma}
\begin{proof}
    For this kind of a system, the output can be represented as a $n$-dimensional vector where the $i^\text{th}$ element can be $1$ if the $i^\text{th}$ document is selected to be in the top-$k$ output and $0$ otherwise. Since the system outputs top-$k$ documents, there would be exactly $k$ elements in the output vector with value $1$ and the remaining $n-k$ elements would be $0$.
    Thus, the output space can be defined as the set of $n$-dimensional vectors with only binary elements such that the element-wise sum remains $k$.
    However, the maximum number of independently varying values in the output space would still be $n-1$ since one has to have the knowledge of $n-1$ elements of the output vector (in the worst case, the certainty about $n-1$ elements can confirm the last element of the vector, i.e., last element will be equal to $k-\{\text{sum of } n-1\text{ known elements}\}$) to fully construct the output vector in all possible scenarios.
    Thus DoFo$($Index-based Search Engine$)=n-1$.
\end{proof}

\begin{definition}
    \textbf{\textit{Knowledge Base Question Answering}}: It is a system that, given a user question (ideally having short answers) and pre-prepared knowledge base (often in the form of knowledge graphs), matches the question with contents of the knowledge base and outputs the answer accordingly.
\end{definition}
\begin{lemma}
    AI systems designed for \textbf{knowledge base question answering} regression have $0$ degree of freedom in output given a user query and the pre-prepared knowledge base. 
\end{lemma}
\begin{proof}
    It is straightforward to notice that given an input question and the pre-prepared knowledge base, the answering system is simply matching the question with the contents of the knowledge base which already contains the answer. Thus, such a system has no freedom in choosing it’s output, when the knowledge base is already provided. DoFo$($Question-Answering$|$Knowledge Base$)=0$
\end{proof}
\begin{definition}
    \textbf{\textit{Rule-based Chatbot}} \cite{adamopoulou2020overview}: It is an interactive system which follows a predefined set of rules to transition between various predefined states. Each state of the chatbot presents a predefined set of options for the user who interacts with it. Based on the selection by the user, the chatbot follows the corresponding associated rule to transition to the next state. These chatbots have been used extensively for automated customer support. 
\end{definition}
\begin{lemma}
    A \textbf{rule-based chatbot} system $0$ degree of freedom in output given the states, options and transition rules. 
\end{lemma}
\begin{proof}
    Since the states and rules completely define how and what would be the output of the chatbot system, the chatbot itself has no freedom in choosing their output. Thus DoFo$($Rule-based Chatbot$)=0$. Note that we assume that the rules do not involve any calls to another AI system with higher DoFo.
\end{proof}

%% file: references.bib
@article{park2024ai,
  title={AI deception: A survey of examples, risks, and potential solutions},
  author={Park, Peter S and Goldstein, Simon and O’Gara, Aidan and Chen, Michael and Hendrycks, Dan},
  journal={Patterns},
  volume={5},
  number={5},
  year={2024},
  publisher={Elsevier}
}

@inproceedings{longpre2023flan,
  title={The flan collection: Designing data and methods for effective instruction tuning},
  author={Longpre, Shayne and Hou, Le and Vu, Tu and Webson, Albert and Chung, Hyung Won and Tay, Yi and Zhou, Denny and Le, Quoc V and Zoph, Barret and Wei, Jason and others},
  booktitle={International Conference on Machine Learning},
  pages={22631--22648},
  year={2023},
  organization={PMLR}
}

@article{kaur2022trustworthy,
  title={Trustworthy artificial intelligence: a review},
  author={Kaur, Davinder and Uslu, Suleyman and Rittichier, Kaley J and Durresi, Arjan},
  journal={ACM computing surveys (CSUR)},
  volume={55},
  number={2},
  pages={1--38},
  year={2022},
  publisher={ACM New York, NY}
}

@article{cheng2021socially,
  title={Socially responsible ai algorithms: Issues, purposes, and challenges},
  author={Cheng, Lu and Varshney, Kush R and Liu, Huan},
  journal={Journal of Artificial Intelligence Research},
  volume={71},
  pages={1137--1181},
  year={2021}
}

@inproceedings{weng2023large,
  title={Large Language Models are Better Reasoners with Self-Verification},
  author={Weng, Yixuan and Zhu, Minjun and Xia, Fei and Li, Bin and He, Shizhu and Liu, Shengping and Sun, Bin and Liu, Kang and Zhao, Jun},
  booktitle={Findings of the Association for Computational Linguistics: EMNLP 2023},
  pages={2550--2575},
  year={2023}
}

@article{guo2025deepseek,
  title={Deepseek-r1: Incentivizing reasoning capability in llms via reinforcement learning},
  author={Guo, Daya and Yang, Dejian and Zhang, Haowei and Song, Junxiao and Zhang, Ruoyu and Xu, Runxin and Zhu, Qihao and Ma, Shirong and Wang, Peiyi and Bi, Xiao and others},
  journal={arXiv preprint arXiv:2501.12948},
  year={2025}
}

@inproceedings{huang2023towards,
  title={Towards Reasoning in Large Language Models: A Survey},
  author={Huang, Jie and Chang, Kevin Chen-Chuan},
  booktitle={Findings of the Association for Computational Linguistics: ACL 2023},
  pages={1049--1065},
  year={2023}
}

@article{sheth2023neurosymbolic,
  title={Neurosymbolic artificial intelligence (why, what, and how)},
  author={Sheth, Amit and Roy, Kaushik and Gaur, Manas},
  journal={IEEE Intelligent Systems},
  volume={38},
  number={3},
  pages={56--62},
  year={2023},
  publisher={IEEE}
}

@article{garcez2023neurosymbolic,
  title={Neurosymbolic AI: The 3 rd wave},
  author={Garcez, Artur d’Avila and Lamb, Luis C},
  journal={Artificial Intelligence Review},
  volume={56},
  number={11},
  pages={12387--12406},
  year={2023},
  publisher={Springer}
}

@article{zhu2024survey,
  title={A Survey on Model Compression for Large Language Models},
  author={Zhu, Xunyu and Li, Jian and Liu, Yong and Ma, Can and Wang, Weiping},
  journal={Transactions of the Association for Computational Linguistics},
  volume={12},
  pages={1556--1577},
  year={2024}
}

@article{yang2024survey,
  title={Survey on knowledge distillation for large language models: methods, evaluation, and application},
  author={Yang, Chuanpeng and Zhu, Yao and Lu, Wang and Wang, Yidong and Chen, Qian and Gao, Chenlong and Yan, Bingjie and Chen, Yiqiang},
  journal={ACM Transactions on Intelligent Systems and Technology},
  year={2024},
  publisher={ACM New York, NY}
}

@inproceedings{shnitzerlarge,
  title={Large Language Model Routing with Benchmark Datasets},
  author={Shnitzer, Tal and Ou, Anthony and Silva, M{\'\i}rian and Soule, Kate and Sun, Yuekai and Solomon, Justin and Thompson, Neil and Yurochkin, Mikhail},
  booktitle={First Conference on Language Modeling}
}

@inproceedings{dong2024building,
  author       = {Yi Dong and
                  Ronghui Mu and
                  Gaojie Jin and
                  Yi Qi and
                  Jinwei Hu and
                  Xingyu Zhao and
                  Jie Meng and
                  Wenjie Ruan and
                  Xiaowei Huang},
  title        = {Position: Building Guardrails for Large Language Models Requires Systematic
                  Design},
  booktitle    = {Forty-first International Conference on Machine Learning, {ICML} 2024,
                  Vienna, Austria, July 21-27, 2024},
  year         = {2024}
}

@article{wu2024avatar,
  title={Avatar: Optimizing llm agents for tool usage via contrastive reasoning},
  author={Wu, Shirley and Zhao, Shiyu and Huang, Qian and Huang, Kexin and Yasunaga, Michihiro and Cao, Kaidi and Ioannidis, Vassilis and Subbian, Karthik and Leskovec, Jure and Zou, James Y},
  journal={Advances in Neural Information Processing Systems},
  volume={37},
  pages={25981--26010},
  year={2024}
}

@inproceedings{li2025review,
  title={A Review of Prominent Paradigms for LLM-Based Agents: Tool Use, Planning (Including RAG), and Feedback Learning},
  author={Li, Xinzhe},
  booktitle={Proceedings of the 31st International Conference on Computational Linguistics},
  pages={9760--9779},
  year={2025}
}

@article{ramnath2025systematic,
  title={A Systematic Survey of Automatic Prompt Optimization Techniques},
  author={Ramnath, Kiran and Zhou, Kang and Guan, Sheng and Mishra, Soumya Smruti and Qi, Xuan and Shen, Zhengyuan and Wang, Shuai and Woo, Sangmin and Jeoung, Sullam and Wang, Yawei and others},
  journal={arXiv preprint arXiv:2502.16923},
  year={2025}
}

@article{sahoo2024systematic,
  title={A systematic survey of prompt engineering in large language models: Techniques and applications},
  author={Sahoo, Pranab and Singh, Ayush Kumar and Saha, Sriparna and Jain, Vinija and Mondal, Samrat and Chadha, Aman},
  journal={arXiv preprint arXiv:2402.07927},
  year={2024}
}

@article{rafailov2023direct,
  title={Direct preference optimization: Your language model is secretly a reward model},
  author={Rafailov, Rafael and Sharma, Archit and Mitchell, Eric and Manning, Christopher D and Ermon, Stefano and Finn, Chelsea},
  journal={Advances in Neural Information Processing Systems},
  volume={36},
  pages={53728--53741},
  year={2023}
}

@article{wang2024comprehensive,
  title={A comprehensive survey of LLM alignment techniques: RLHF, RLAIF, PPO, DPO and more},
  author={Wang, Zhichao and Bi, Bin and Pentyala, Shiva Kumar and Ramnath, Kiran and Chaudhuri, Sougata and Mehrotra, Shubham and Mao, Xiang-Bo and Asur, Sitaram and others},
  journal={arXiv preprint arXiv:2407.16216},
  year={2024}
}

@article{wach2023dark,
  title={The dark side of generative artificial intelligence: A critical analysis of controversies and risks of ChatGPT},
  author={Wach, Krzysztof and Duong, Cong Doanh and Ejdys, Joanna and Kazlauskait{\.e}, R{\=u}ta and Korzynski, Pawel and Mazurek, Grzegorz and Paliszkiewicz, Joanna and Ziemba, Ewa},
  journal={Entrepreneurial Business and Economics Review},
  volume={11},
  number={2},
  pages={7--30},
  year={2023}
}

@article{rillig2023risks,
  title={Risks and benefits of large language models for the environment},
  author={Rillig, Matthias C and {\AA}gerstrand, Marlene and Bi, Mohan and Gould, Kenneth A and Sauerland, Uli},
  journal={Environmental science \& technology},
  volume={57},
  number={9},
  pages={3464--3466},
  year={2023},
  publisher={ACS Publications}
}

@article{fei2022towards,
  title={Towards artificial general intelligence via a multimodal foundation model},
  author={Fei, Nanyi and Lu, Zhiwu and Gao, Yizhao and Yang, Guoxing and Huo, Yuqi and Wen, Jingyuan and Lu, Haoyu and Song, Ruihua and Gao, Xin and Xiang, Tao and others},
  journal={Nature Communications},
  volume={13},
  number={1},
  pages={3094},
  year={2022},
  publisher={Nature Publishing Group UK London}
}

@article{goktas2025shaping,
  title={Shaping the future of healthcare: Ethical clinical challenges and pathways to trustworthy AI},
  author={Goktas, Polat and Grzybowski, Andrzej},
  journal={Journal of Clinical Medicine},
  volume={14},
  number={5},
  pages={1605},
  year={2025},
  publisher={MDPI}
}

@article{zhu2023large,
  title={Large language models for information retrieval: A survey},
  author={Zhu, Yutao and Yuan, Huaying and Wang, Shuting and Liu, Jiongnan and Liu, Wenhan and Deng, Chenlong and Chen, Haonan and Liu, Zheng and Dou, Zhicheng and Wen, Ji-Rong},
  journal={arXiv preprint arXiv:2308.07107},
  year={2023}
}

@inproceedings{liang2024survey,
  title={A Survey of Multimodel Large Language Models},
  author={Liang, Zijing and Xu, Yanjie and Hong, Yifan and Shang, Penghui and Wang, Qi and Fu, Qiang and Liu, Ke},
  booktitle={Proceedings of the 3rd International Conference on Computer, Artificial Intelligence and Control Engineering},
  pages={405--409},
  year={2024}
}

@article{chang2024survey,
  title={A survey on evaluation of large language models},
  author={Chang, Yupeng and Wang, Xu and Wang, Jindong and Wu, Yuan and Yang, Linyi and Zhu, Kaijie and Chen, Hao and Yi, Xiaoyuan and Wang, Cunxiang and Wang, Yidong and others},
  journal={ACM transactions on intelligent systems and technology},
  volume={15},
  number={3},
  pages={1--45},
  year={2024},
  publisher={ACM New York, NY}
}

@article{wei2022emergent,
  author       = {Jason Wei and
                  Yi Tay and
                  Rishi Bommasani and
                  Colin Raffel and
                  Barret Zoph and
                  Sebastian Borgeaud and
                  Dani Yogatama and
                  Maarten Bosma and
                  Denny Zhou and
                  Donald Metzler and
                  Ed H. Chi and
                  Tatsunori Hashimoto and
                  Oriol Vinyals and
                  Percy Liang and
                  Jeff Dean and
                  William Fedus},
  title        = {Emergent Abilities of Large Language Models},
  journal      = {Trans. Mach. Learn. Res.},
  volume       = {2022},
  year         = {2022}
}

@inproceedings{li2024large,
  title={Large Language Models for Generative Recommendation: A Survey and Visionary Discussions},
  author={Li, Lei and Zhang, Yongfeng and Liu, Dugang and Chen, Li},
  booktitle={Proceedings of the 2024 Joint International Conference on Computational Linguistics, Language Resources and Evaluation (LREC-COLING 2024)},
  pages={10146--10159},
  year={2024}
}

@article{croitoru2023diffusion,
  title={Diffusion models in vision: A survey},
  author={Croitoru, Florinel-Alin and Hondru, Vlad and Ionescu, Radu Tudor and Shah, Mubarak},
  journal={IEEE Transactions on Pattern Analysis and Machine Intelligence},
  volume={45},
  number={9},
  pages={10850--10869},
  year={2023},
  publisher={IEEE}
}

@article{yang2023diffusion,
  title={Diffusion models: A comprehensive survey of methods and applications},
  author={Yang, Ling and Zhang, Zhilong and Song, Yang and Hong, Shenda and Xu, Runsheng and Zhao, Yue and Zhang, Wentao and Cui, Bin and Yang, Ming-Hsuan},
  journal={ACM Computing Surveys},
  volume={56},
  number={4},
  pages={1--39},
  year={2023},
  publisher={ACM New York, NY, USA}
}

@article{gozalo2023survey,
  title={A survey of Generative AI Applications},
  author={Gozalo-Brizuela, Roberto and Garrido-Merch{\'a}n, Eduardo C},
  journal={arXiv preprint arXiv:2306.02781},
  year={2023}
}

@article{dotan2024evolving,
  author       = {Ravit Dotan and
                  Borhane Blili{-}Hamelin and
                  Ravi Madhavan and
                  Jeanna Matthews and
                  Joshua Scarpino},
  title        = {Evolving {AI} Risk Management: {A} Maturity Model based on the {NIST}
                  {AI} Risk Management Framework},
  journal      = {CoRR},
  volume       = {abs/2401.15229},
  year         = {2024},
  url          = {https://doi.org/10.48550/arXiv.2401.15229},
  doi          = {10.48550/ARXIV.2401.15229}
}

@article{clarke2019regulatory,
  title={Regulatory alternatives for AI},
  author={Clarke, Roger},
  journal={Computer Law \& Security Review},
  volume={35},
  number={4},
  pages={398--409},
  year={2019},
  publisher={Elsevier}
}

@inproceedings{schiff2020s,
  title={What's next for ai ethics, policy, and governance? a global overview},
  author={Schiff, Daniel and Biddle, Justin and Borenstein, Jason and Laas, Kelly},
  booktitle={Proceedings of the AAAI/ACM Conference on AI, Ethics, and Society},
  pages={153--158},
  year={2020}
}

@article{taeihagh2021governance,
  title={Governance of artificial intelligence},
  author={Taeihagh, Araz},
  journal={Policy and society},
  volume={40},
  number={2},
  pages={137--157},
  year={2021},
  publisher={Oxford University Press}
}

@article{dafoe2018ai,
  title={AI governance: a research agenda},
  author={Dafoe, Allan},
  journal={Governance of AI Program, Future of Humanity Institute, University of Oxford: Oxford, UK},
  volume={1442},
  pages={1443},
  year={2018}
}

@article{batool2025ai,
  title={AI governance: a systematic literature review},
  author={Batool, Amna and Zowghi, Didar and Bano, Muneera},
  journal={AI and Ethics},
  pages={1--15},
  year={2025},
  publisher={Springer}
}

@article{liang2022advances,
  title={Advances, challenges and opportunities in creating data for trustworthy AI},
  author={Liang, Weixin and Tadesse, Girmaw Abebe and Ho, Daniel and Fei-Fei, Li and Zaharia, Matei and Zhang, Ce and Zou, James},
  journal={Nature Machine Intelligence},
  volume={4},
  number={8},
  pages={669--677},
  year={2022},
  publisher={Nature Publishing Group UK London}
}

@article{ganguly2023review,
  title={A review of the role of causality in developing trustworthy ai systems},
  author={Ganguly, Niloy and Fazlija, Dren and Badar, Maryam and Fisichella, Marco and Sikdar, Sandipan and Schrader, Johanna and Wallat, Jonas and Rudra, Koustav and Koubarakis, Manolis and Patro, Gourab K and others},
  journal={arXiv preprint arXiv:2302.06975},
  year={2023}
}

@article{nishant2020artificial,
  title={Artificial intelligence for sustainability: Challenges, opportunities, and a research agenda},
  author={Nishant, Rohit and Kennedy, Mike and Corbett, Jacqueline},
  journal={International journal of information management},
  volume={53},
  pages={102104},
  year={2020},
  publisher={Elsevier}
}

@article{van2021sustainable,
  title={Sustainable AI: AI for sustainability and the sustainability of AI},
  author={Van Wynsberghe, Aimee},
  journal={AI and Ethics},
  volume={1},
  number={3},
  pages={213--218},
  year={2021},
  publisher={Springer}
}

@article{liu2024survey,
  title={A survey on hallucination in large vision-language models},
  author={Liu, Hanchao and Xue, Wenyuan and Chen, Yifei and Chen, Dapeng and Zhao, Xiutian and Wang, Ke and Hou, Liping and Li, Rongjun and Peng, Wei},
  journal={arXiv preprint arXiv:2402.00253},
  year={2024}
}

@article{mou2024sg,
  title={SG-Bench: Evaluating LLM Safety Generalization Across Diverse Tasks and Prompt Types},
  author={Mou, Yutao and Zhang, Shikun and Ye, Wei},
  journal={Advances in Neural Information Processing Systems},
  volume={37},
  pages={123032--123054},
  year={2024}
}

@article{salhab2024systematic,
  title={A systematic literature review on ai safety: Identifying trends, challenges and future directions},
  author={Salhab, Wissam and Ameyed, Darine and Jaafar, Fehmi and Mcheick, Hamid},
  journal={IEEE Access},
  year={2024},
  publisher={IEEE}
}

@article{akhtar2021advances,
  title={Advances in adversarial attacks and defenses in computer vision: A survey},
  author={Akhtar, Naveed and Mian, Ajmal and Kardan, Navid and Shah, Mubarak},
  journal={IEEE Access},
  volume={9},
  pages={155161--155196},
  year={2021},
  publisher={IEEE}
}

@article{silva2020opportunities,
  title={Opportunities and challenges in deep learning adversarial robustness: A survey},
  author={Silva, Samuel Henrique and Najafirad, Peyman},
  journal={arXiv preprint arXiv:2007.00753},
  year={2020}
}

@inproceedings{wang2022measure,
  title={Measure and Improve Robustness in NLP Models: A Survey},
  author={Wang, Xuezhi and Wang, Haohan and Yang, Diyi},
  booktitle={Proceedings of the 2022 Conference of the North American Chapter of the Association for Computational Linguistics: Human Language Technologies},
  pages={4569--4586},
  year={2022}
}

@article{tocchetti2025ai,
  title={Ai robustness: a human-centered perspective on technological challenges and opportunities},
  author={Tocchetti, Andrea and Corti, Lorenzo and Balayn, Agathe and Yurrita, Mireia and Lippmann, Philip and Brambilla, Marco and Yang, Jie},
  journal={ACM Computing Surveys},
  volume={57},
  number={6},
  pages={1--38},
  year={2025},
  publisher={ACM New York, NY}
}

@article{bartl2025gender,
  title={Gender bias in natural language processing and computer vision: A comparative survey},
  author={Bartl, Marion and Mandal, Abhishek and Leavy, Susan and Little, Suzanne},
  journal={ACM Computing Surveys},
  volume={57},
  number={6},
  pages={1--36},
  year={2025},
  publisher={ACM New York, NY}
}

@article{zhang2018visual,
  title={Visual interpretability for deep learning: a survey},
  author={Zhang, Quan-shi and Zhu, Song-Chun},
  journal={Frontiers of Information Technology \& Electronic Engineering},
  volume={19},
  number={1},
  pages={27--39},
  year={2018},
  publisher={Springer}
}

@article{carvalho2019machine,
  title={Machine learning interpretability: A survey on methods and metrics},
  author={Carvalho, Diogo V and Pereira, Eduardo M and Cardoso, Jaime S},
  journal={Electronics},
  volume={8},
  number={8},
  pages={832},
  year={2019},
  publisher={Multidisciplinary Digital Publishing Institute}
}

@article{das2020opportunities,
  title={Opportunities and challenges in explainable artificial intelligence (xai): A survey},
  author={Das, Arun and Rad, Paul},
  journal={arXiv preprint arXiv:2006.11371},
  year={2020}
}

@article{zhao2025visual,
  title={Visual content privacy protection: A survey},
  author={Zhao, Ruoyu and Zhang, Yushu and Wang, Tao and Wen, Wenying and Xiang, Yong and Cao, Xiaochun},
  journal={ACM Computing Surveys},
  volume={57},
  number={5},
  pages={1--36},
  year={2025},
  publisher={ACM New York, NY}
}

@article{das2025security,
  title={Security and privacy challenges of large language models: A survey},
  author={Das, Badhan Chandra and Amini, M Hadi and Wu, Yanzhao},
  journal={ACM Computing Surveys},
  volume={57},
  number={6},
  pages={1--39},
  year={2025},
  publisher={ACM New York, NY}
}

@article{yao2024survey,
  title={A survey on large language model (llm) security and privacy: The good, the bad, and the ugly},
  author={Yao, Yifan and Duan, Jinhao and Xu, Kaidi and Cai, Yuanfang and Sun, Zhibo and Zhang, Yue},
  journal={High-Confidence Computing},
  pages={100211},
  year={2024},
  publisher={Elsevier}
}

@article{rigaki2023survey,
  title={A survey of privacy attacks in machine learning},
  author={Rigaki, Maria and Garcia, Sebastian},
  journal={ACM Computing Surveys},
  volume={56},
  number={4},
  pages={1--34},
  year={2023},
  publisher={ACM New York, NY}
}

@article{liu2021machine,
  title={When machine learning meets privacy: A survey and outlook},
  author={Liu, Bo and Ding, Ming and Shaham, Sina and Rahayu, Wenny and Farokhi, Farhad and Lin, Zihuai},
  journal={ACM Computing Surveys (CSUR)},
  volume={54},
  number={2},
  pages={1--36},
  year={2021},
  publisher={ACM New York, NY, USA}
}

@article{parraga2025fairness,
  title={Fairness in Deep Learning: A survey on vision and language research},
  author={Parraga, Otavio and More, Martin D and Oliveira, Christian M and Gavenski, Nathan S and Kupssinsk{\"u}, Lucas S and Medronha, Adilson and Moura, Luis V and Sim{\~o}es, Gabriel S and Barros, Rodrigo C},
  journal={ACM Computing Surveys},
  volume={57},
  number={6},
  pages={1--40},
  year={2025},
  publisher={ACM New York, NY}
}

@inproceedings{adamopoulou2020overview,
  title={An overview of chatbot technology},
  author={Adamopoulou, Eleni and Moussiades, Lefteris},
  booktitle={Artificial Intelligence Applications and Innovations: 16th IFIP WG 12.5 International Conference, AIAI 2020, Neos Marmaras, Greece, June 5--7, 2020, Proceedings, Part II 16},
  pages={373--383},
  year={2020},
  organization={Springer}
}

@inproceedings{moratanch2017survey,
  title={A survey on extractive text summarization},
  author={Moratanch, N and Chitrakala, S},
  booktitle={2017 international conference on computer, communication and signal processing (ICCCSP)},
  pages={1--6},
  year={2017},
  organization={IEEE}
}

@article{nguyen2024llms,
  title={Llms for generating and evaluating counterfactuals: A comprehensive study},
  author={Nguyen, Van Bach and Youssef, Paul and Seifert, Christin and Schl{\"o}tterer, J{\"o}rg},
  journal={arXiv preprint arXiv:2405.00722},
  year={2024}
}

@article{chuang2024faithlm,
  title={FaithLM: Towards faithful explanations for large language models},
  author={Chuang, Yu-Neng and Wang, Guanchu and Chang, Chia-Yuan and Tang, Ruixiang and Zhong, Shaochen and Yang, Fan and Du, Mengnan and Cai, Xuanting and Hu, Xia},
  journal={arXiv preprint arXiv:2402.04678},
  year={2024}
}

@article{matton2025walk,
  title={Walk the Talk? Measuring the Faithfulness of Large Language Model Explanations},
  author={Matton, Katie and Ness, Robert Osazuwa and Guttag, John and K{\i}c{\i}man, Emre},
  journal={arXiv preprint arXiv:2504.14150},
  year={2025}
}

@article{huang2023can,
  title={Can large language models explain themselves? a study of llm-generated self-explanations},
  author={Huang, Shiyuan and Mamidanna, Siddarth and Jangam, Shreedhar and Zhou, Yilun and Gilpin, Leilani H},
  journal={arXiv preprint arXiv:2310.11207},
  year={2023}
}

@inproceedings{perez2023discovering,
  title={Discovering language model behaviors with model-written evaluations},
  author={Perez, Ethan and Ringer, Sam and Lukosiute, Kamile and Nguyen, Karina and Chen, Edwin and Heiner, Scott and Pettit, Craig and Olsson, Catherine and Kundu, Sandipan and Kadavath, Saurav and others},
  booktitle={Findings of the Association for Computational Linguistics: ACL 2023},
  pages={13387--13434},
  year={2023}
}

@article{turpin2023language,
  title={Language models don't always say what they think: Unfaithful explanations in chain-of-thought prompting},
  author={Turpin, Miles and Michael, Julian and Perez, Ethan and Bowman, Samuel},
  journal={Advances in Neural Information Processing Systems},
  volume={36},
  pages={74952--74965},
  year={2023}
}

@inproceedings{atanasova2023faithfulness,
  title={Faithfulness Tests for Natural Language Explanations},
  author={Atanasova, Pepa and Camburu, Oana-Maria and Lioma, Christina and Lukasiewicz, Thomas and Simonsen, Jakob Grue and Augenstein, Isabelle},
  booktitle={Proceedings of the 61st Annual Meeting of the Association for Computational Linguistics (Volume 2: Short Papers)},
  pages={283--294},
  year={2023}
}

@article{lanham2023measuring,
  title={Measuring faithfulness in chain-of-thought reasoning},
  author={Lanham, Tamera and Chen, Anna and Radhakrishnan, Ansh and Steiner, Benoit and Denison, Carson and Hernandez, Danny and Li, Dustin and Durmus, Esin and Hubinger, Evan and Kernion, Jackson and others},
  journal={arXiv preprint arXiv:2307.13702},
  year={2023}
}

@inproceedings{jacovi2020towards,
  title={Towards Faithfully Interpretable NLP Systems: How Should We Define and Evaluate Faithfulness?},
  author={Jacovi, Alon and Goldberg, Yoav},
  booktitle={Proceedings of the 58th Annual Meeting of the Association for Computational Linguistics},
  pages={4198--4205},
  year={2020}
}

@article{ye2022unreliability,
  title={The unreliability of explanations in few-shot prompting for textual reasoning},
  author={Ye, Xi and Durrett, Greg},
  journal={Advances in neural information processing systems},
  volume={35},
  pages={30378--30392},
  year={2022}
}

@inproceedings{parcalabescu2024measuring,
  title={On Measuring Faithfulness or Self-consistency of Natural Language Explanations},
  author={Parcalabescu, Letitia and Frank, Anette},
  booktitle={Proceedings of the 62nd Annual Meeting of the Association for Computational Linguistics (Volume 1: Long Papers)},
  pages={6048--6089},
  year={2024}
}

@article{agarwal2024faithfulness,
  title={Faithfulness vs. plausibility: On the (un) reliability of explanations from large language models},
  author={Agarwal, Chirag and Tanneru, Sree Harsha and Lakkaraju, Himabindu},
  journal={arXiv preprint arXiv:2402.04614},
  year={2024}
}

@inproceedings{madsen2024self,
  title={Are self-explanations from Large Language Models faithful?},
  author={Madsen, Andreas and Chandar, Sarath and Reddy, Siva},
  booktitle={Findings of the Association for Computational Linguistics ACL 2024},
  pages={295--337},
  year={2024}
}

@inproceedings{chen2024models,
  title={Do Models Explain Themselves? Counterfactual Simulatability of Natural Language Explanations},
  author={Chen, Yanda and Zhong, Ruiqi and Ri, Narutatsu and Zhao, Chen and He, He and Steinhardt, Jacob and Yu, Zhou and Mckeown, Kathleen},
  booktitle={International Conference on Machine Learning},
  pages={7880--7904},
  year={2024},
  organization={PMLR}
}

@inproceedings{kokalj2021bert,
  title={BERT meets shapley: Extending SHAP explanations to transformer-based classifiers},
  author={Kokalj, Enja and {\v{S}}krlj, Bla{\v{z}} and Lavra{\v{c}}, Nada and Pollak, Senja and Robnik-{\v{S}}ikonja, Marko},
  booktitle={Proceedings of the EACL hackashop on news media content analysis and automated report generation},
  pages={16--21},
  year={2021}
}

@article{chen2023algorithms,
  title={Algorithms to estimate Shapley value feature attributions},
  author={Chen, Hugh and Covert, Ian C and Lundberg, Scott M and Lee, Su-In},
  journal={Nature Machine Intelligence},
  volume={5},
  number={6},
  pages={590--601},
  year={2023},
  publisher={Nature Publishing Group UK London}
}

@article{wei2022chain,
  title={Chain-of-thought prompting elicits reasoning in large language models},
  author={Wei, Jason and Wang, Xuezhi and Schuurmans, Dale and Bosma, Maarten and Xia, Fei and Chi, Ed and Le, Quoc V and Zhou, Denny and others},
  journal={Advances in neural information processing systems},
  volume={35},
  pages={24824--24837},
  year={2022}
}

@article{yeh2018representer,
  title={Representer point selection for explaining deep neural networks},
  author={Yeh, Chih-Kuan and Kim, Joon and Yen, Ian En-Hsu and Ravikumar, Pradeep K},
  journal={Advances in neural information processing systems},
  volume={31},
  year={2018}
}

@inproceedings{wu2021polyjuice,
  title={Polyjuice: Generating Counterfactuals for Explaining, Evaluating, and Improving Models},
  author={Wu, Tongshuang and Ribeiro, Marco Tulio and Heer, Jeffrey and Weld, Daniel S},
  booktitle={Proceedings of the 59th Annual Meeting of the Association for Computational Linguistics and the 11th International Joint Conference on Natural Language Processing (Volume 1: Long Papers)},
  pages={6707--6723},
  year={2021}
}

@inproceedings{li2021contextualized,
  title={Contextualized Perturbation for Textual Adversarial Attack},
  author={Li, Dianqi and Zhang, Yizhe and Peng, Hao and Chen, Liqun and Brockett, Chris and Sun, Ming-Ting and Dolan, William B},
  booktitle={Proceedings of the 2021 Conference of the North American Chapter of the Association for Computational Linguistics: Human Language Technologies},
  pages={5053--5069},
  year={2021}
}

@article{derose2020attention,
  title={Attention flows: Analyzing and comparing attention mechanisms in language models},
  author={DeRose, Joseph F and Wang, Jiayao and Berger, Matthew},
  journal={IEEE Transactions on Visualization and Computer Graphics},
  volume={27},
  number={2},
  pages={1160--1170},
  year={2020},
  publisher={IEEE}
}

@inproceedings{hoover2020exbert,
  title={exBERT: A Visual Analysis Tool to Explore Learned Representations in Transformer Models},
  author={Hoover, Benjamin and Strobelt, Hendrik and Gehrmann, Sebastian},
  booktitle={Proceedings of the 58th Annual Meeting of the Association for Computational Linguistics: System Demonstrations},
  pages={187--196},
  year={2020}
}

@inproceedings{voita2019analyzing,
  title={Analyzing Multi-Head Self-Attention: Specialized Heads Do the Heavy Lifting, the Rest Can Be Pruned},
  author={Voita, Elena and Talbot, David and Moiseev, Fedor and Sennrich, Rico and Titov, Ivan},
  booktitle={Proceedings of the 57th Annual Meeting of the Association for Computational Linguistics},
  pages={5797--5808},
  year={2019}
}

@article{montavon2019layer,
  title={Layer-wise relevance propagation: an overview},
  author={Montavon, Gr{\'e}goire and Binder, Alexander and Lapuschkin, Sebastian and Samek, Wojciech and M{\"u}ller, Klaus-Robert},
  journal={Explainable AI: interpreting, explaining and visualizing deep learning},
  pages={193--209},
  year={2019},
  publisher={Springer}
}

@inproceedings{du2019attribution,
  title={On attribution of recurrent neural network predictions via additive decomposition},
  author={Du, Mengnan and Liu, Ninghao and Yang, Fan and Ji, Shuiwang and Hu, Xia},
  booktitle={The world wide web conference},
  pages={383--393},
  year={2019}
}

@inproceedings{ribeiro2016should,
  title={" Why should i trust you?" Explaining the predictions of any classifier},
  author={Ribeiro, Marco Tulio and Singh, Sameer and Guestrin, Carlos},
  booktitle={Proceedings of the 22nd ACM SIGKDD international conference on knowledge discovery and data mining},
  pages={1135--1144},
  year={2016}
}

@inproceedings{sundararajan2017axiomatic,
  title={Axiomatic attribution for deep networks},
  author={Sundararajan, Mukund and Taly, Ankur and Yan, Qiqi},
  booktitle={International conference on machine learning},
  pages={3319--3328},
  year={2017},
  organization={PMLR}
}

@article{ancona2019gradient,
  title={Gradient-based attribution methods},
  author={Ancona, Marco and Ceolini, Enea and {\"O}ztireli, Cengiz and Gross, Markus},
  journal={Explainable AI: Interpreting, explaining and visualizing deep learning},
  pages={169--191},
  year={2019},
  publisher={Springer}
}

@article{wu2020perturbed,
  title={Perturbed masking: Parameter-free probing for analyzing and interpreting BERT},
  author={Wu, Zhiyong and Chen, Yun and Kao, Ben and Liu, Qun},
  journal={arXiv preprint arXiv:2004.14786},
  year={2020}
}

@article{ivanovs2021perturbation,
  title={Perturbation-based methods for explaining deep neural networks: A survey},
  author={Ivanovs, Maksims and Kadikis, Roberts and Ozols, Kaspars},
  journal={Pattern Recognition Letters},
  volume={150},
  pages={228--234},
  year={2021},
  publisher={Elsevier}
}

@inproceedings{ghorbani2019data,
  title={Data shapley: Equitable valuation of data for machine learning},
  author={Ghorbani, Amirata and Zou, James},
  booktitle={International conference on machine learning},
  pages={2242--2251},
  year={2019},
  organization={PMLR}
}

@article{pruthi2020estimating,
  title={Estimating training data influence by tracing gradient descent},
  author={Pruthi, Garima and Liu, Frederick and Kale, Satyen and Sundararajan, Mukund},
  journal={Advances in Neural Information Processing Systems},
  volume={33},
  pages={19920--19930},
  year={2020}
}

@article{dabkowski2017real,
  title={Real time image saliency for black box classifiers},
  author={Dabkowski, Piotr and Gal, Yarin},
  journal={Advances in neural information processing systems},
  volume={30},
  year={2017}
}

@article{lundberg2017unified,
  title={A unified approach to interpreting model predictions},
  author={Lundberg, Scott M and Lee, Su-In},
  journal={Advances in neural information processing systems},
  volume={30},
  year={2017}
}

@article{zhang2025building,
  title={Building Bridges, Not Walls--Advancing Interpretability by Unifying Feature, Data, and Model Component Attribution},
  author={Zhang, Shichang and Han, Tessa and Bhalla, Usha and Lakkaraju, Himabindu},
  journal={arXiv preprint arXiv:2501.18887},
  year={2025}
}

@inproceedings{palit2023towards,
  title={Towards vision-language mechanistic interpretability: A causal tracing tool for blip},
  author={Palit, Vedant and Pandey, Rohan and Arora, Aryaman and Liang, Paul Pu},
  booktitle={Proceedings of the IEEE/CVF International Conference on Computer Vision},
  pages={2856--2861},
  year={2023}
}

@article{maini2024llm,
  title={LLM Dataset Inference: Did you train on my dataset?},
  author={Maini, Pratyush and Jia, Hengrui and Papernot, Nicolas and Dziedzic, Adam},
  journal={Advances in Neural Information Processing Systems},
  volume={37},
  pages={124069--124092},
  year={2024}
}

@article{conmy2023towards,
  title={Towards automated circuit discovery for mechanistic interpretability},
  author={Conmy, Arthur and Mavor-Parker, Augustine and Lynch, Aengus and Heimersheim, Stefan and Garriga-Alonso, Adri{\`a}},
  journal={Advances in Neural Information Processing Systems},
  volume={36},
  pages={16318--16352},
  year={2023}
}

@article{doshi2017towards,
  title={Towards a rigorous science of interpretable machine learning},
  author={Doshi-Velez, Finale and Kim, Been},
  journal={arXiv preprint arXiv:1702.08608},
  year={2017}
}

@article{burkart2021survey,
  title={A survey on the explainability of supervised machine learning},
  author={Burkart, Nadia and Huber, Marco F},
  journal={Journal of Artificial Intelligence Research},
  volume={70},
  pages={245--317},
  year={2021}
}

@article{zhao2024explainability,
  title={Explainability for large language models: A survey},
  author={Zhao, Haiyan and Chen, Hanjie and Yang, Fan and Liu, Ninghao and Deng, Huiqi and Cai, Hengyi and Wang, Shuaiqiang and Yin, Dawei and Du, Mengnan},
  journal={ACM Transactions on Intelligent Systems and Technology},
  volume={15},
  number={2},
  pages={1--38},
  year={2024},
  publisher={ACM New York, NY}
}

@article{liu2025rethinking,
  title={Rethinking machine unlearning for large language models},
  author={Liu, Sijia and Yao, Yuanshun and Jia, Jinghan and Casper, Stephen and Baracaldo, Nathalie and Hase, Peter and Yao, Yuguang and Liu, Chris Yuhao and Xu, Xiaojun and Li, Hang and others},
  journal={Nature Machine Intelligence},
  pages={1--14},
  year={2025},
  publisher={Nature Publishing Group UK London}
}

@inproceedings{finocchiaro2021bridging,
  title={Bridging machine learning and mechanism design towards algorithmic fairness},
  author={Finocchiaro, Jessie and Maio, Roland and Monachou, Faidra and Patro, Gourab K and Raghavan, Manish and Stoica, Ana-Andreea and Tsirtsis, Stratis},
  booktitle={Proceedings of the 2021 ACM conference on fairness, accountability, and transparency},
  pages={489--503},
  year={2021}
}

@inproceedings{lees2022new,
  title={A new generation of perspective api: Efficient multilingual character-level transformers},
  author={Lees, Alyssa and Tran, Vinh Q and Tay, Yi and Sorensen, Jeffrey and Gupta, Jai and Metzler, Donald and Vasserman, Lucy},
  booktitle={Proceedings of the 28th ACM SIGKDD conference on knowledge discovery and data mining},
  pages={3197--3207},
  year={2022}
}

@article{cao2024nothing,
  title={Nothing in excess: Mitigating the exaggerated safety for llms via safety-conscious activation steering},
  author={Cao, Zouying and Yang, Yifei and Zhao, Hai},
  journal={arXiv preprint arXiv:2408.11491},
  year={2024}
}

@inproceedings{rottger2024xstest,
  title={XSTest: A Test Suite for Identifying Exaggerated Safety Behaviours in Large Language Models},
  author={R{\"o}ttger, Paul and Kirk, Hannah and Vidgen, Bertie and Attanasio, Giuseppe and Bianchi, Federico and Hovy, Dirk},
  booktitle={Proceedings of the 2024 Conference of the North American Chapter of the Association for Computational Linguistics: Human Language Technologies (Volume 1: Long Papers)},
  pages={5377--5400},
  year={2024}
}

@article{chou2023villandiffusion,
  title={Villandiffusion: A unified backdoor attack framework for diffusion models},
  author={Chou, Sheng-Yen and Chen, Pin-Yu and Ho, Tsung-Yi},
  journal={Advances in Neural Information Processing Systems},
  volume={36},
  pages={33912--33964},
  year={2023}
}

@article{gao2020backdoor,
  title={Backdoor attacks and countermeasures on deep learning: A comprehensive review},
  author={Gao, Yansong and Doan, Bao Gia and Zhang, Zhi and Ma, Siqi and Zhang, Jiliang and Fu, Anmin and Nepal, Surya and Kim, Hyoungshick},
  journal={arXiv preprint arXiv:2007.10760},
  year={2020}
}

@article{udeshi2022model,
  title={Model agnostic defence against backdoor attacks in machine learning},
  author={Udeshi, Sakshi and Peng, Shanshan and Woo, Gerald and Loh, Lionell and Rawshan, Louth and Chattopadhyay, Sudipta},
  journal={IEEE Transactions on Reliability},
  volume={71},
  number={2},
  pages={880--895},
  year={2022},
  publisher={IEEE}
}

@article{zhang2024backdoor,
  title={Backdoor attacks and defenses targeting multi-domain ai models: A comprehensive review},
  author={Zhang, Shaobo and Pan, Yimeng and Liu, Qin and Yan, Zheng and Choo, Kim-Kwang Raymond and Wang, Guojun},
  journal={ACM Computing Surveys},
  volume={57},
  number={4},
  pages={1--35},
  year={2024},
  publisher={ACM New York, NY}
}

@inproceedings{salem2022dynamic,
  title={Dynamic backdoor attacks against machine learning models},
  author={Salem, Ahmed and Wen, Rui and Backes, Michael and Ma, Shiqing and Zhang, Yang},
  booktitle={2022 IEEE 7th European Symposium on Security and Privacy (EuroS\&P)},
  pages={703--718},
  year={2022},
  organization={IEEE}
}

@inproceedings{gehman2020realtoxicityprompts,
  title={RealToxicityPrompts: Evaluating Neural Toxic Degeneration in Language Models},
  author={Gehman, Samuel and Gururangan, Suchin and Sap, Maarten and Choi, Yejin and Smith, Noah A},
  booktitle={Findings of the Association for Computational Linguistics: EMNLP 2020},
  pages={3356--3369},
  year={2020}
}

@misc{perspective_api,
  title = {Perspective API},
year={2022},
author={Perspective-API},
  howpublished = {\url{https://perspectiveapi.com/}},
  note = {Accessed: 2025-03-24}
}

@article{mohseni2022taxonomy,
  title={Taxonomy of machine learning safety: A survey and primer},
  author={Mohseni, Sina and Wang, Haotao and Xiao, Chaowei and Yu, Zhiding and Wang, Zhangyang and Yadawa, Jay},
  journal={ACM Computing Surveys},
  volume={55},
  number={8},
  pages={1--38},
  year={2022},
  publisher={ACM New York, NY}
}

@article{amodei2016concrete,
  title={Concrete problems in AI safety},
  author={Amodei, Dario and Olah, Chris and Steinhardt, Jacob and Christiano, Paul and Schulman, John and Man{\'e}, Dan},
  journal={arXiv preprint arXiv:1606.06565},
  year={2016}
}

@article{wei2023jailbroken,
  title={Jailbroken: How does llm safety training fail?},
  author={Wei, Alexander and Haghtalab, Nika and Steinhardt, Jacob},
  journal={Advances in Neural Information Processing Systems},
  volume={36},
  pages={80079--80110},
  year={2023}
}

@article{ji2023beavertails,
  title={Beavertails: Towards improved safety alignment of llm via a human-preference dataset},
  author={Ji, Jiaming and Liu, Mickel and Dai, Josef and Pan, Xuehai and Zhang, Chi and Bian, Ce and Chen, Boyuan and Sun, Ruiyang and Wang, Yizhou and Yang, Yaodong},
  journal={Advances in Neural Information Processing Systems},
  volume={36},
  pages={24678--24704},
  year={2023}
}

@article{rebedea2023nemo,
  title={Nemo guardrails: A toolkit for controllable and safe llm applications with programmable rails},
  author={Rebedea, Traian and Dinu, Razvan and Sreedhar, Makesh and Parisien, Christopher and Cohen, Jonathan},
  journal={arXiv preprint arXiv:2310.10501},
  year={2023}
}

@inproceedings{wen2023unveiling,
  title={Unveiling the Implicit Toxicity in Large Language Models},
  author={Wen, Jiaxin and Ke, Pei and Sun, Hao and Zhang, Zhexin and Li, Chengfei and Bai, Jinfeng and Huang, Minlie},
  booktitle={Proceedings of the 2023 Conference on Empirical Methods in Natural Language Processing},
  pages={1322--1338},
  year={2023}
}

@inproceedings{shen2024anything,
  title={" do anything now": Characterizing and evaluating in-the-wild jailbreak prompts on large language models},
  author={Shen, Xinyue and Chen, Zeyuan and Backes, Michael and Shen, Yun and Zhang, Yang},
  booktitle={Proceedings of the 2024 on ACM SIGSAC Conference on Computer and Communications Security},
  pages={1671--1685},
  year={2024}
}

@inproceedings{poe2024conflict,
  title={The Conflict Between Algorithmic Fairness and Non-Discrimination: An Analysis of Fair Automated Hiring},
  author={Poe, Robert Lee and El Mestari, Soumia Zohra},
  booktitle={Proceedings of the 2024 ACM Conference on Fairness, Accountability, and Transparency},
  pages={1907--1916},
  year={2024}
}

@inproceedings{giannopoulos2024fairness,
  title={Fairness in AI: challenges in bridging the gap between algorithms and law},
  author={Giannopoulos, Giorgos and Psalla, Maria and Kavouras, Loukas and Sacharidis, Dimitris and Marecek, Jakub and Matilla, Germ{\'a}n M and Emiris, Ioannis},
  booktitle={2024 IEEE 40th International Conference on Data Engineering Workshops (ICDEW)},
  pages={217--225},
  year={2024},
  organization={IEEE}
}

@inproceedings{kong2022intersectionally,
  title={Are “intersectionally fair” ai algorithms really fair to women of color? a philosophical analysis},
  author={Kong, Youjin},
  booktitle={Proceedings of the 2022 ACM Conference on Fairness, Accountability, and Transparency},
  pages={485--494},
  year={2022}
}

@article{mehrabi2021survey,
  title={A survey on bias and fairness in machine learning},
  author={Mehrabi, Ninareh and Morstatter, Fred and Saxena, Nripsuta and Lerman, Kristina and Galstyan, Aram},
  journal={ACM computing surveys (CSUR)},
  volume={54},
  number={6},
  pages={1--35},
  year={2021},
  publisher={ACM New York, NY, USA}
}

@article{ferrara2023fairness,
  title={Fairness and bias in artificial intelligence: A brief survey of sources, impacts, and mitigation strategies},
  author={Ferrara, Emilio},
  journal={Sci},
  volume={6},
  number={1},
  pages={3},
  year={2023},
  publisher={MDPI}
}

@article{sun2024trustllm,
  title={Trustllm: Trustworthiness in large language models},
  author={Sun, Lichao and Huang, Yue and Wang, Haoran and Wu, Siyuan and Zhang, Qihui and Gao, Chujie and Huang, Yixin and Lyu, Wenhan and Zhang, Yixuan and Li, Xiner and others},
  journal={arXiv preprint arXiv:2401.05561},
  volume={3},
  year={2024}
}

@inproceedings{sreelekha2018statistical,
  title={Statistical vs. rule-based machine translation: A comparative study on indian languages},
  author={Sreelekha, S and Bhattacharyya, Pushpak and Malathi, D},
  booktitle={International Conference on Intelligent Computing and Applications: ICICA 2016},
  pages={663--675},
  year={2018},
  organization={Springer}
}

@article{torregrosa2020aspects,
  title={Aspects of terminological and named entity knowledge within rule-based machine translation models for under-resourced neural machine translation scenarios},
  author={Torregrosa, Daniel and Pasricha, Nivranshu and Masoud, Maraim and Chakravarthi, Bharathi Raja and Alonso, Juan and Casas, Noe and Arcan, Mihael},
  journal={arXiv preprint arXiv:2009.13398},
  year={2020}
}

@article{zhu2023multilingual,
  title={Multilingual machine translation with large language models: Empirical results and analysis},
  author={Zhu, Wenhao and Liu, Hongyi and Dong, Qingxiu and Xu, Jingjing and Huang, Shujian and Kong, Lingpeng and Chen, Jiajun and Li, Lei},
  journal={arXiv preprint arXiv:2304.04675},
  year={2023}
}

@article{xu2023paradigm,
  title={A paradigm shift in machine translation: Boosting translation performance of large language models},
  author={Xu, Haoran and Kim, Young Jin and Sharaf, Amr and Awadalla, Hany Hassan},
  journal={arXiv preprint arXiv:2309.11674},
  year={2023}
}

@inproceedings{mvechura2022taxonomy,
  title={A taxonomy of bias-causing ambiguities in machine translation},
  author={M{\v{e}}chura, Michal},
  booktitle={Proceedings of the 4th Workshop on Gender Bias in Natural Language Processing (GeBNLP)},
  pages={168--173},
  year={2022}
}

@article{wang2022generalizing,
  title={Generalizing to unseen domains: A survey on domain generalization},
  author={Wang, Jindong and Lan, Cuiling and Liu, Chang and Ouyang, Yidong and Qin, Tao and Lu, Wang and Chen, Yiqiang and Zeng, Wenjun and Yu, Philip S},
  journal={IEEE transactions on knowledge and data engineering},
  volume={35},
  number={8},
  pages={8052--8072},
  year={2022},
  publisher={IEEE}
}

@article{drenkow2021systematic,
  title={A systematic review of robustness in deep learning for computer vision: Mind the gap?},
  author={Drenkow, Nathan and Sani, Numair and Shpitser, Ilya and Unberath, Mathias},
  journal={arXiv preprint arXiv:2112.00639},
  year={2021}
}

@article{liu2023trustworthy,
  title={Trustworthy llms: a survey and guideline for evaluating large language models' alignment},
  author={Liu, Yang and Yao, Yuanshun and Ton, Jean-Francois and Zhang, Xiaoying and Guo, Ruocheng and Cheng, Hao and Klochkov, Yegor and Taufiq, Muhammad Faaiz and Li, Hang},
  journal={arXiv preprint arXiv:2308.05374},
  year={2023}
}

@article{hendrycks2019benchmarking,
  title={Benchmarking neural network robustness to common corruptions and perturbations},
  author={Hendrycks, Dan and Dietterich, Thomas},
  journal={arXiv preprint arXiv:1903.12261},
  year={2019}
}

@inproceedings{ghosh2024generative,
  title={Do generative AI models output harm while representing non-Western cultures: Evidence from a community-centered approach},
  author={Ghosh, Sourojit and Venkit, Pranav Narayanan and Gautam, Sanjana and Wilson, Shomir and Caliskan, Aylin},
  booktitle={Proceedings of the AAAI/ACM Conference on AI, Ethics, and Society},
  volume={7},
  pages={476--489},
  year={2024}
}

@article{cortiz2020ethical,
  title={Ethical and technical challenges of AI in tackling hate speech},
  author={Cortiz, Diogo and Zubiaga, Arkaitz},
  journal={The International Review of Information Ethics},
  volume={29},
  year={2020}
}

@incollection{rughinis2024generative,
  title={Generative AI and social engines of hate},
  author={Rughiniș, R{\u{a}}zvan and Rughiniș, Cosima and Bran, Emanuela},
  booktitle={Regulating Hate Speech Created by Generative AI},
  pages={1--18},
  year={2024},
  publisher={Auerbach Publications}
}

@article{zhang2023survey,
  title={A survey on complex factual question answering},
  author={Zhang, Lingxi and Zhang, Jing and Ke, Xirui and Li, Haoyang and Huang, Xinmei and Shao, Zhonghui and Cao, Shulin and Lv, Xin},
  journal={AI Open},
  volume={4},
  pages={1--12},
  year={2023},
  publisher={Elsevier}
}

@article{gallegos2024bias,
  title={Bias and fairness in large language models: A survey},
  author={Gallegos, Isabel O and Rossi, Ryan A and Barrow, Joe and Tanjim, Md Mehrab and Kim, Sungchul and Dernoncourt, Franck and Yu, Tong and Zhang, Ruiyi and Ahmed, Nesreen K},
  journal={Computational Linguistics},
  volume={50},
  number={3},
  pages={1097--1179},
  year={2024},
  publisher={MIT Press 255 Main Street, 9th Floor, Cambridge, Massachusetts 02142, USA~…}
}

@inproceedings{dorn2024harmful,
  title={Harmful speech detection by language models exhibits gender-queer dialect bias},
  author={Dorn, Rebecca and Kezar, Lee and Morstatter, Fred and Lerman, Kristina},
  booktitle={Proceedings of the 4th ACM Conference on Equity and Access in Algorithms, Mechanisms, and Optimization},
  pages={1--12},
  year={2024}
}

@article{anagnostopoulos2022biased,
  title={Biased opinion dynamics: when the devil is in the details},
  author={Anagnostopoulos, Aris and Becchetti, Luca and Cruciani, Emilio and Pasquale, Francesco and Rizzo, Sara},
  journal={Information Sciences},
  volume={593},
  pages={49--63},
  year={2022},
  publisher={Elsevier}
}

@article{huang2024bias,
  title={Bias in Opinion Summarisation from Pre-training to Adaptation: A Case Study in Political Bias},
  author={Huang, Nannan and Fayek, Haytham and Zhang, Xiuzhen},
  journal={arXiv preprint arXiv:2402.00322},
  year={2024}
}

@article{shakil2024abstractive,
  title={Abstractive text summarization: State of the art, challenges, and improvements},
  author={Shakil, Hassan and Farooq, Ahmad and Kalita, Jugal},
  journal={Neurocomputing},
  pages={128255},
  year={2024},
  publisher={Elsevier}
}

@inproceedings{wang2023decodingtrust,
  title={DecodingTrust: A Comprehensive Assessment of Trustworthiness in GPT Models.},
  author={Wang, Boxin and Chen, Weixin and Pei, Hengzhi and Xie, Chulin and Kang, Mintong and Zhang, Chenhui and Xu, Chejian and Xiong, Zidi and Dutta, Ritik and Schaeffer, Rylan and others},
  booktitle={NeurIPS},
  year={2023}
}

@article{mitra2018introduction,
  title={An introduction to neural information retrieval},
  author={Mitra, Bhaskar and Craswell, Nick and others},
  journal={Foundations and Trends{\textregistered} in Information Retrieval},
  volume={13},
  number={1},
  pages={1--126},
  year={2018},
  publisher={Now Publishers, Inc.}
}

@article{mallen2022not,
  title={When not to trust language models: Investigating effectiveness of parametric and non-parametric memories},
  author={Mallen, Alex and Asai, Akari and Zhong, Victor and Das, Rajarshi and Khashabi, Daniel and Hajishirzi, Hannaneh},
  journal={arXiv preprint arXiv:2212.10511},
  year={2022}
}

@article{ji2023survey,
  title={Survey of hallucination in natural language generation},
  author={Ji, Ziwei and Lee, Nayeon and Frieske, Rita and Yu, Tiezheng and Su, Dan and Xu, Yan and Ishii, Etsuko and Bang, Ye Jin and Madotto, Andrea and Fung, Pascale},
  journal={ACM computing surveys},
  volume={55},
  number={12},
  pages={1--38},
  year={2023},
  publisher={ACM New York, NY}
}

@article{huang2025survey,
  title={A survey on hallucination in large language models: Principles, taxonomy, challenges, and open questions},
  author={Huang, Lei and Yu, Weijiang and Ma, Weitao and Zhong, Weihong and Feng, Zhangyin and Wang, Haotian and Chen, Qianglong and Peng, Weihua and Feng, Xiaocheng and Qin, Bing and others},
  journal={ACM Transactions on Information Systems},
  volume={43},
  number={2},
  pages={1--55},
  year={2025},
  publisher={ACM New York, NY}
}

@article{ouyang2022training,
  title={Training language models to follow instructions with human feedback},
  author={Ouyang, Long and Wu, Jeffrey and Jiang, Xu and Almeida, Diogo and Wainwright, Carroll and Mishkin, Pamela and Zhang, Chong and Agarwal, Sandhini and Slama, Katarina and Ray, Alex and others},
  journal={Advances in neural information processing systems},
  volume={35},
  pages={27730--27744},
  year={2022}
}

@inproceedings{gal2016dropout,
  title={Dropout as a bayesian approximation: Representing model uncertainty in deep learning},
  author={Gal, Yarin and Ghahramani, Zoubin},
  booktitle={international conference on machine learning},
  pages={1050--1059},
  year={2016},
  organization={PMLR}
}

@article{diwan2021multilingual,
  title={Multilingual and code-switching ASR challenges for low resource Indian languages},
  author={Diwan, Anuj and Vaideeswaran, Rakesh and Shah, Sanket and Singh, Ankita and Raghavan, Srinivasa and Khare, Shreya and Unni, Vinit and Vyas, Saurabh and Rajpuria, Akash and Yarra, Chiranjeevi and others},
  journal={arXiv preprint arXiv:2104.00235},
  year={2021}
}

@inproceedings{huzaifah2024evaluating,
  title={Evaluating code-switching translation with large language models},
  author={Huzaifah, Muhammad and Zheng, Weihua and Chanpaisit, Nattapol and Wu, Kui},
  booktitle={Proceedings of the 2024 Joint International Conference on Computational Linguistics, Language Resources and Evaluation (LREC-COLING 2024)},
  pages={6381--6394},
  year={2024}
}

@article{winata2021multilingual,
  title={Are multilingual models effective in code-switching?},
  author={Winata, Genta Indra and Cahyawijaya, Samuel and Liu, Zihan and Lin, Zhaojiang and Madotto, Andrea and Fung, Pascale},
  journal={arXiv preprint arXiv:2103.13309},
  year={2021}
}

@article{joshi2020state,
  title={The state and fate of linguistic diversity and inclusion in the NLP world},
  author={Joshi, Pratik and Santy, Sebastin and Budhiraja, Amar and Bali, Kalika and Choudhury, Monojit},
  journal={arXiv preprint arXiv:2004.09095},
  year={2020}
}

@article{koehn2017six,
  title={Six challenges for neural machine translation},
  author={Koehn, Philipp and Knowles, Rebecca},
  journal={arXiv preprint arXiv:1706.03872},
  year={2017}
}

@inproceedings{martindale2019identifying,
  title={Identifying fluently inadequate output in neural and statistical machine translation},
  author={Martindale, Marianna and Carpuat, Marine and Duh, Kevin and McNamee, Paul},
  booktitle={Proceedings of Machine Translation Summit XVII: Research Track},
  pages={233--243},
  year={2019}
}

@article{anastasopoulos2019pushing,
  title={Pushing the limits of low-resource morphological inflection},
  author={Anastasopoulos, Antonios and Neubig, Graham},
  journal={arXiv preprint arXiv:1908.05838},
  year={2019}
}

@article{zhang2024code,
  title={Code-mixed LLM: Improve Large Language Models' Capability to Handle Code-Mixing through Reinforcement Learning from AI Feedback},
  author={Zhang, Wenbo and Majumdar, Aditya and Yadav, Amulya},
  journal={arXiv preprint arXiv:2411.09073},
  year={2024}
}

@article{alhanai2024bridging,
  title={Bridging the Gap: Enhancing LLM Performance for Low-Resource African Languages with New Benchmarks, Fine-Tuning, and Cultural Adjustments},
  author={Alhanai, Tuka and Kasumovic, Adam and Ghassemi, Mohammad and Zitzelberger, Aven and Lundin, Jessica and Chabot-Couture, Guillaume},
  journal={arXiv preprint arXiv:2412.12417},
  year={2024}
}

@article{villalobos2022will,
  title={Will we run out of data? Limits of LLM scaling based on human-generated data},
  author={Villalobos, Pablo and Ho, Anson and Sevilla, Jaime and Besiroglu, Tamay and Heim, Lennart and Hobbhahn, Marius},
  journal={arXiv preprint arXiv:2211.04325},
  year={2022}
}

@article{arora2021types,
  title={Types of out-of-distribution texts and how to detect them},
  author={Arora, Udit and Huang, William and He, He},
  journal={arXiv preprint arXiv:2109.06827},
  year={2021}
}

@inproceedings{lin2018mining,
  title={Mining cross-cultural differences and similarities in social media},
  author={Lin, Bill Yuchen and Xu, Frank F and Zhu, Kenny and Hwang, Seung-won},
  booktitle={Proceedings of the 56th Annual Meeting of the Association for Computational Linguistics (Volume 1: Long Papers)},
  pages={709--719},
  year={2018}
}

@article{brown2020language,
  title={Language models are few-shot learners},
  author={Brown, Tom and Mann, Benjamin and Ryder, Nick and Subbiah, Melanie and Kaplan, Jared D and Dhariwal, Prafulla and Neelakantan, Arvind and Shyam, Pranav and Sastry, Girish and Askell, Amanda and others},
  journal={Advances in neural information processing systems},
  volume={33},
  pages={1877--1901},
  year={2020}
}

@article{huang2022generspeech,
  title={Generspeech: Towards style transfer for generalizable out-of-domain text-to-speech},
  author={Huang, Rongjie and Ren, Yi and Liu, Jinglin and Cui, Chenye and Zhao, Zhou},
  journal={Advances in Neural Information Processing Systems},
  volume={35},
  pages={10970--10983},
  year={2022}
}

@inproceedings{gatys2016image,
  title={Image style transfer using convolutional neural networks},
  author={Gatys, Leon A and Ecker, Alexander S and Bethge, Matthias},
  booktitle={Proceedings of the IEEE conference on computer vision and pattern recognition},
  pages={2414--2423},
  year={2016}
}

@article{kanwal2022identifying,
  title={Identifying the evidence of speech emotional dialects using artificial intelligence: A cross-cultural study},
  author={Kanwal, Sofia and Asghar, Sohail and Hussain, Akhtar and Rafique, Adnan},
  journal={Plos one},
  volume={17},
  number={3},
  pages={e0265199},
  year={2022},
  publisher={Public Library of Science San Francisco, CA USA}
}

@article{kapoor2024large,
  title={Large Language Models Must Be Taught to Know What They Don't Know},
  author={Kapoor, Sanyam and Gruver, Nate and Roberts, Manley and Collins, Katherine and Pal, Arka and Bhatt, Umang and Weller, Adrian and Dooley, Samuel and Goldblum, Micah and Wilson, Andrew Gordon},
  journal={arXiv preprint arXiv:2406.08391},
  year={2024}
}

@article{kuhn2023semantic,
  title={Semantic uncertainty: Linguistic invariances for uncertainty estimation in natural language generation},
  author={Kuhn, Lorenz and Gal, Yarin and Farquhar, Sebastian},
  journal={arXiv preprint arXiv:2302.09664},
  year={2023}
}

@article{ulmer2024calibrating,
  title={Calibrating large language models using their generations only},
  author={Ulmer, Dennis and Gubri, Martin and Lee, Hwaran and Yun, Sangdoo and Oh, Seong Joon},
  journal={arXiv preprint arXiv:2403.05973},
  year={2024}
}

@article{prates2020assessing,
  title={Assessing gender bias in machine translation: a case study with google translate},
  author={Prates, Marcelo OR and Avelar, Pedro H and Lamb, Lu{\'\i}s C},
  journal={Neural Computing and Applications},
  volume={32},
  pages={6363--6381},
  year={2020},
  publisher={Springer}
}

@article{ai2023artificial,
  title={Artificial intelligence risk management framework (NIST AI RMF 1.0)},
  author={NIST},
  journal={URL: https://nvlpubs. nist. gov/nistpubs/ai/nist. ai},
  pages={100--1},
  year={2023}
}

@article{king2024rethinking,
  title={Rethinking privacy in the AI era: Policy provocations for a data-centric world},
  author={King, Jennifer and Meinhardt, Caroline},
  journal={Stanford Institute for Human-Centered Artificial Intelligence},
  year={2024}
}

@article{duan2024uncovering,
  title={Uncovering Latent Memories: Assessing Data Leakage and Memorization Patterns in Frontier AI Models},
  author={Duan, Sunny and Khona, Mikail and Iyer, Abhiram and Schaeffer, Rylan and Fiete, Ila R},
  journal={arXiv preprint arXiv:2406.14549},
  year={2024}
}

@article{wei2024memorization,
  title={Memorization in deep learning: A survey},
  author={Wei, Jiaheng and Zhang, Yanjun and Zhang, Leo Yu and Ding, Ming and Chen, Chao and Ong, Kok-Leong and Zhang, Jun and Xiang, Yang},
  journal={arXiv preprint arXiv:2406.03880},
  year={2024}
}

@inproceedings{mireshghallah2022empirical,
  title={An empirical analysis of memorization in fine-tuned autoregressive language models},
  author={Mireshghallah, Fatemehsadat and Uniyal, Archit and Wang, Tianhao and Evans, David K and Berg-Kirkpatrick, Taylor},
  booktitle={Proceedings of the 2022 Conference on Empirical Methods in Natural Language Processing},
  pages={1816--1826},
  year={2022}
}

@inproceedings{carlini2019secret,
  title={The secret sharer: Evaluating and testing unintended memorization in neural networks},
  author={Carlini, Nicholas and Liu, Chang and Erlingsson, {\'U}lfar and Kos, Jernej and Song, Dawn},
  booktitle={28th USENIX security symposium (USENIX security 19)},
  pages={267--284},
  year={2019}
}

@article{del2023bounding,
  title={Bounding information leakage in machine learning},
  author={Del Grosso, Ganesh and Pichler, Georg and Palamidessi, Catuscia and Piantanida, Pablo},
  journal={Neurocomputing},
  volume={534},
  pages={1--17},
  year={2023},
  publisher={Elsevier}
}

@article{chen2024multi,
  title={A multi-perspective analysis of memorization in large language models},
  author={Chen, Bowen and Han, Namgi and Miyao, Yusuke},
  journal={arXiv preprint arXiv:2405.11577},
  year={2024}
}

@inproceedings{carlini2022quantifying,
  title={Quantifying memorization across neural language models},
  author={Carlini, Nicholas and Ippolito, Daphne and Jagielski, Matthew and Lee, Katherine and Tramer, Florian and Zhang, Chiyuan},
  booktitle={The Eleventh International Conference on Learning Representations},
  year={2022}
}

@article{ippolito2022preventing,
  title={Preventing verbatim memorization in language models gives a false sense of privacy},
  author={Ippolito, Daphne and Tram{\`e}r, Florian and Nasr, Milad and Zhang, Chiyuan and Jagielski, Matthew and Lee, Katherine and Choquette-Choo, Christopher A and Carlini, Nicholas},
  journal={arXiv preprint arXiv:2210.17546},
  year={2022}
}

@article{huang2022large,
  title={Are large pre-trained language models leaking your personal information?},
  author={Huang, Jie and Shao, Hanyin and Chang, Kevin Chen-Chuan},
  journal={arXiv preprint arXiv:2205.12628},
  year={2022}
}

@inproceedings{thomas2020investigating,
  title={Investigating the impact of pre-trained word embeddings on memorization in neural networks},
  author={Thomas, Aleena and Adelani, David Ifeoluwa and Davody, Ali and Mogadala, Aditya and Klakow, Dietrich},
  booktitle={Text, Speech, and Dialogue: 23rd International Conference, TSD 2020, Brno, Czech Republic, September 8--11, 2020, Proceedings 23},
  pages={273--281},
  year={2020},
  organization={Springer}
}

@article{desai2024between,
  title={Between copyright and computer science: The law and ethics of generative ai},
  author={Desai, Deven R and Riedl, Mark},
  journal={Nw. J. Tech. \& Intell. Prop.},
  volume={22},
  pages={55},
  year={2024},
  publisher={HeinOnline}
}

@article{demir2023ai,
  title={AI-Induced Copyright Infringement: Application of Legal Neutrality Based on the Qualitative Similarity Criterion},
  author={Demir, Mehmet Salih},
  journal={Available at SSRN 4689391},
  year={2023}
}

@article{henderson2023foundation,
  title={Foundation models and fair use},
  author={Henderson, Peter and Li, Xuechen and Jurafsky, Dan and Hashimoto, Tatsunori and Lemley, Mark A and Liang, Percy},
  journal={Journal of Machine Learning Research},
  volume={24},
  number={400},
  pages={1--79},
  year={2023}
}

@article{hartmann2023sok,
  title={Sok: Memorization in general-purpose large language models},
  author={Hartmann, Valentin and Suri, Anshuman and Bindschaedler, Vincent and Evans, David and Tople, Shruti and West, Robert},
  journal={arXiv preprint arXiv:2310.18362},
  year={2023}
}

@inproceedings{rahman2023beyond,
  title={Beyond fair use: Legal risk evaluation for training LLMs on copyrighted text},
  author={Rahman, Noorjahan and Santacana, Eduardo},
  booktitle={ICML Workshop on Generative AI and Law},
  year={2023}
}

@article{cooper2024files,
  title={The Files are in the Computer: Copyright, Memorization, and Generative AI},
  author={Cooper, A Feder and Grimmelmann, James},
  journal={arXiv preprint arXiv:2404.12590},
  year={2024}
}

@article{brittain2023getty,
  title={Getty Images lawsuit says Stability AI misused photos to train AI},
  author={Brittain, Blake},
  journal={Reuters. Accessed: Jun},
  volume={5},
  year={2023}
}

@article{grynbaum2023times,
  title={The Times sues OpenAI and Microsoft over AI use of copyrighted work},
  author={Grynbaum, Michael M and Mac, Ryan},
  journal={The New York Times},
  volume={27},
  year={2023}
}

@article{evans2021truthful,
  title={Truthful AI: Developing and governing AI that does not lie},
  author={Evans, Owain and Cotton-Barratt, Owen and Finnveden, Lukas and Bales, Adam and Balwit, Avital and Wills, Peter and Righetti, Luca and Saunders, William},
  journal={arXiv preprint arXiv:2110.06674},
  year={2021}
}

@article{lin2021truthfulqa,
  title={Truthfulqa: Measuring how models mimic human falsehoods},
  author={Lin, Stephanie and Hilton, Jacob and Evans, Owain},
  journal={arXiv preprint arXiv:2109.07958},
  year={2021}
}

@article{rawte2023survey,
  title={A survey of hallucination in large foundation models},
  author={Rawte, Vipula and Sheth, Amit and Das, Amitava},
  journal={arXiv preprint arXiv:2309.05922},
  year={2023}
}

@article{sriramanan2024llm,
  title={Llm-check: Investigating detection of hallucinations in large language models},
  author={Sriramanan, Gaurang and Bharti, Siddhant and Sadasivan, Vinu Sankar and Saha, Shoumik and Kattakinda, Priyatham and Feizi, Soheil},
  journal={Advances in Neural Information Processing Systems},
  volume={37},
  pages={34188--34216},
  year={2024}
}

@article{ye2023cognitive,
  title={Cognitive mirage: A review of hallucinations in large language models},
  author={Ye, Hongbin and Liu, Tong and Zhang, Aijia and Hua, Wei and Jia, Weiqiang},
  journal={arXiv preprint arXiv:2309.06794},
  year={2023}
}

@article{zhang2023siren,
  title={Siren's song in the AI ocean: a survey on hallucination in large language models},
  author={Zhang, Yue and Li, Yafu and Cui, Leyang and Cai, Deng and Liu, Lemao and Fu, Tingchen and Huang, Xinting and Zhao, Enbo and Zhang, Yu and Chen, Yulong and others},
  journal={arXiv preprint arXiv:2309.01219},
  year={2023}
}

@article{liu2024exploring,
  title={Exploring and evaluating hallucinations in llm-powered code generation},
  author={Liu, Fang and Liu, Yang and Shi, Lin and Huang, Houkun and Wang, Ruifeng and Yang, Zhen and Zhang, Li and Li, Zhongqi and Ma, Yuchi},
  journal={arXiv preprint arXiv:2404.00971},
  year={2024}
}

@article{zhu2024halueval,
  title={Halueval-wild: Evaluating hallucinations of language models in the wild},
  author={Zhu, Zhiying and Yang, Yiming and Sun, Zhiqing},
  journal={arXiv preprint arXiv:2403.04307},
  year={2024}
}

@article{ayyamperumal2024current,
  title={Current state of LLM Risks and AI Guardrails},
  author={Ayyamperumal, Suriya Ganesh and Ge, Limin},
  journal={arXiv preprint arXiv:2406.12934},
  year={2024}
}

@article{pal2023med,
  title={Med-halt: Medical domain hallucination test for large language models},
  author={Pal, Ankit and Umapathi, Logesh Kumar and Sankarasubbu, Malaikannan},
  journal={arXiv preprint arXiv:2307.15343},
  year={2023}
}

@article{barman2024dark,
  title={The dark side of language models: Exploring the potential of llms in multimedia disinformation generation and dissemination},
  author={Barman, Dipto and Guo, Ziyi and Conlan, Owen},
  journal={Machine Learning with Applications},
  pages={100545},
  year={2024},
  publisher={Elsevier}
}

@article{chen2023can,
  title={Can llm-generated misinformation be detected?},
  author={Chen, Canyu and Shu, Kai},
  journal={arXiv preprint arXiv:2309.13788},
  year={2023}
}

@article{manakul2023selfcheckgpt,
  title={Selfcheckgpt: Zero-resource black-box hallucination detection for generative large language models},
  author={Manakul, Potsawee and Liusie, Adian and Gales, Mark JF},
  journal={arXiv preprint arXiv:2303.08896},
  year={2023}
}

@article{chen2024inside,
  title={INSIDE: LLMs' internal states retain the power of hallucination detection},
  author={Chen, Chao and Liu, Kai and Chen, Ze and Gu, Yi and Wu, Yue and Tao, Mingyuan and Fu, Zhihang and Ye, Jieping},
  journal={arXiv preprint arXiv:2402.03744},
  year={2024}
}

@article{wang2022self,
  title={Self-consistency improves chain of thought reasoning in language models},
  author={Wang, Xuezhi and Wei, Jason and Schuurmans, Dale and Le, Quoc and Chi, Ed and Narang, Sharan and Chowdhery, Aakanksha and Zhou, Denny},
  journal={arXiv preprint arXiv:2203.11171},
  year={2022}
}

@article{lewis2020retrieval,
  title={Retrieval-augmented generation for knowledge-intensive nlp tasks},
  author={Lewis, Patrick and Perez, Ethan and Piktus, Aleksandra and Petroni, Fabio and Karpukhin, Vladimir and Goyal, Naman and K{\"u}ttler, Heinrich and Lewis, Mike and Yih, Wen-tau and Rockt{\"a}schel, Tim and others},
  journal={Advances in neural information processing systems},
  volume={33},
  pages={9459--9474},
  year={2020}
}

@article{niu2023ragtruth,
  title={Ragtruth: A hallucination corpus for developing trustworthy retrieval-augmented language models},
  author={Niu, Cheng and Wu, Yuanhao and Zhu, Juno and Xu, Siliang and Shum, Kashun and Zhong, Randy and Song, Juntong and Zhang, Tong},
  journal={arXiv preprint arXiv:2401.00396},
  year={2023}
}

@article{varshney2023stitch,
  title={A stitch in time saves nine: Detecting and mitigating hallucinations of llms by validating low-confidence generation},
  author={Varshney, Neeraj and Yao, Wenlin and Zhang, Hongming and Chen, Jianshu and Yu, Dong},
  journal={arXiv preprint arXiv:2307.03987},
  year={2023}
}

@article{sharma2023towards,
  title={Towards understanding sycophancy in language models},
  author={Sharma, Mrinank and Tong, Meg and Korbak, Tomasz and Duvenaud, David and Askell, Amanda and Bowman, Samuel R and Cheng, Newton and Durmus, Esin and Hatfield-Dodds, Zac and Johnston, Scott R and others},
  journal={arXiv preprint arXiv:2310.13548},
  year={2023}
}

@article{wei2023simple,
  title={Simple synthetic data reduces sycophancy in large language models},
  author={Wei, Jerry and Huang, Da and Lu, Yifeng and Zhou, Denny and Le, Quoc V},
  journal={arXiv preprint arXiv:2308.03958},
  year={2023}
}

@article{liu2025truth,
  title={TRUTH DECAY: Quantifying Multi-Turn Sycophancy in Language Models},
  author={Liu, Joshua and Jain, Aarav and Takuri, Soham and Vege, Srihan and Akalin, Aslihan and Zhu, Kevin and O'Brien, Sean and Sharma, Vasu},
  journal={arXiv preprint arXiv:2503.11656},
  year={2025}
}

@article{chern2024behonest,
  title={BeHonest: Benchmarking Honesty in Large Language Models},
  author={Chern, Steffi and Hu, Zhulin and Yang, Yuqing and Chern, Ethan and Guo, Yuan and Jin, Jiahe and Wang, Binjie and Liu, Pengfei},
  journal={arXiv preprint arXiv:2406.13261},
  year={2024}
}

@article{zhao2025aligning,
  title={Aligning large language models for faithful integrity against opposing argument},
  author={Zhao, Yong and Deng, Yang and Ng, See-Kiong and Chua, Tat-Seng},
  journal={arXiv preprint arXiv:2501.01336},
  year={2025}
}

@article{wang2023can,
  title={Can ChatGPT defend its belief in truth? evaluating LLM reasoning via debate},
  author={Wang, Boshi and Yue, Xiang and Sun, Huan},
  journal={arXiv preprint arXiv:2305.13160},
  year={2023}
}

@article{bai2022training,
  title={Training a helpful and harmless assistant with reinforcement learning from human feedback},
  author={Bai, Yuntao and Jones, Andy and Ndousse, Kamal and Askell, Amanda and Chen, Anna and DasSarma, Nova and Drain, Dawn and Fort, Stanislav and Ganguli, Deep and Henighan, Tom and others},
  journal={arXiv preprint arXiv:2204.05862},
  year={2022}
}

@article{cotra2021ai,
  title={Why AI alignment could be hard with modern deep learning},
  author={Cotra, Ajeya},
  journal={Cold Takes},
  year={2021}
}

@article{wang2023mint,
  title={Mint: Evaluating llms in multi-turn interaction with tools and language feedback},
  author={Wang, Xingyao and Wang, Zihan and Liu, Jiateng and Chen, Yangyi and Yuan, Lifan and Peng, Hao and Ji, Heng},
  journal={arXiv preprint arXiv:2309.10691},
  year={2023}
}

@article{bowman2022measuring,
  title={Measuring progress on scalable oversight for large language models},
  author={Bowman, Samuel R and Hyun, Jeeyoon and Perez, Ethan and Chen, Edwin and Pettit, Craig and Heiner, Scott and Luko{\v{s}}i{\=u}t{\.e}, Kamil{\.e} and Askell, Amanda and Jones, Andy and Chen, Anna and others},
  journal={arXiv preprint arXiv:2211.03540},
  year={2022}
}

@inproceedings{liu2024formalizing,
  title={Formalizing and benchmarking prompt injection attacks and defenses},
  author={Liu, Yupei and Jia, Yuqi and Geng, Runpeng and Jia, Jinyuan and Gong, Neil Zhenqiang},
  booktitle={33rd USENIX Security Symposium (USENIX Security 24)},
  pages={1831--1847},
  year={2024}
}

@inproceedings{greshake2023not,
  title={Not what you've signed up for: Compromising real-world llm-integrated applications with indirect prompt injection},
  author={Greshake, Kai and Abdelnabi, Sahar and Mishra, Shailesh and Endres, Christoph and Holz, Thorsten and Fritz, Mario},
  booktitle={Proceedings of the 16th ACM Workshop on Artificial Intelligence and Security},
  pages={79--90},
  year={2023}
}

@article{wang2023safeguarding,
  title={Safeguarding crowdsourcing surveys from ChatGPT with prompt injection},
  author={Wang, Chaofan and Freire, Samuel Kernan and Zhang, Mo and Wei, Jing and Goncalves, Jorge and Kostakos, Vassilis and Sarsenbayeva, Zhanna and Schneegass, Christina and Bozzon, Alessandro and Niforatos, Evangelos},
  journal={arXiv preprint arXiv:2306.08833},
  year={2023}
}

@article{liu2024automatic,
  title={Automatic and universal prompt injection attacks against large language models},
  author={Liu, Xiaogeng and Yu, Zhiyuan and Zhang, Yizhe and Zhang, Ning and Xiao, Chaowei},
  journal={arXiv preprint arXiv:2403.04957},
  year={2024}
}

@inproceedings{yip2023novel,
  title={A novel evaluation framework for assessing resilience against prompt injection attacks in large language models},
  author={Yip, Daniel Wankit and Esmradi, Aysan and Chan, Chun Fai},
  booktitle={2023 IEEE Asia-Pacific Conference on Computer Science and Data Engineering (CSDE)},
  pages={1--5},
  year={2023},
  organization={IEEE}
}

@article{perez2022ignore,
  title={Ignore previous prompt: Attack techniques for language models},
  author={Perez, F{\'a}bio and Ribeiro, Ian},
  journal={arXiv preprint arXiv:2211.09527},
  year={2022}
}

@article{kaddour2023challenges,
  title={Challenges and applications of large language models},
  author={Kaddour, Jean and Harris, Joshua and Mozes, Maximilian and Bradley, Herbie and Raileanu, Roberta and McHardy, Robert},
  journal={arXiv preprint arXiv:2307.10169},
  year={2023}
}

@article{li2023multi,
  title={Multi-step jailbreaking privacy attacks on chatgpt},
  author={Li, Haoran and Guo, Dadi and Fan, Wei and Xu, Mingshi and Huang, Jie and Meng, Fanpu and Song, Yangqiu},
  journal={arXiv preprint arXiv:2304.05197},
  year={2023}
}

@article{zou2023universal,
  title={Universal and transferable adversarial attacks on aligned language models},
  author={Zou, Andy and Wang, Zifan and Carlini, Nicholas and Nasr, Milad and Kolter, J Zico and Fredrikson, Matt},
  journal={arXiv preprint arXiv:2307.15043},
  year={2023}
}

@article{haim2022reconstructing,
  title={Reconstructing training data from trained neural networks},
  author={Haim, Niv and Vardi, Gal and Yehudai, Gilad and Shamir, Ohad and Irani, Michal},
  journal={Advances in Neural Information Processing Systems},
  volume={35},
  pages={22911--22924},
  year={2022}
}

@article{runkel2024training,
  title={Training Data Reconstruction: Privacy due to Uncertainty?},
  author={Runkel, Christina and Gandikota, Kanchana Vaishnavi and Geiping, Jonas and Sch{\"o}nlieb, Carola-Bibiane and Moeller, Michael},
  journal={arXiv preprint arXiv:2412.08544},
  year={2024}
}

@inproceedings{balle2022reconstructing,
  title={Reconstructing training data with informed adversaries},
  author={Balle, Borja and Cherubin, Giovanni and Hayes, Jamie},
  booktitle={2022 IEEE Symposium on Security and Privacy (SP)},
  pages={1138--1156},
  year={2022},
  organization={IEEE}
}

@article{zhu2019deep,
  title={Deep leakage from gradients},
  author={Zhu, Ligeng and Liu, Zhijian and Han, Song},
  journal={Advances in neural information processing systems},
  volume={32},
  year={2019}
}

@article{zhao2020idlg,
  title={idlg: Improved deep leakage from gradients},
  author={Zhao, Bo and Mopuri, Konda Reddy and Bilen, Hakan},
  journal={arXiv preprint arXiv:2001.02610},
  year={2020}
}

@inproceedings{fredrikson2015model,
  title={Model inversion attacks that exploit confidence information and basic countermeasures},
  author={Fredrikson, Matt and Jha, Somesh and Ristenpart, Thomas},
  booktitle={Proceedings of the 22nd ACM SIGSAC conference on computer and communications security},
  pages={1322--1333},
  year={2015}
}

@inproceedings{carlini2021extracting,
  title={Extracting training data from large language models},
  author={Carlini, Nicholas and Tramer, Florian and Wallace, Eric and Jagielski, Matthew and Herbert-Voss, Ariel and Lee, Katherine and Roberts, Adam and Brown, Tom and Song, Dawn and Erlingsson, Ulfar and others},
  booktitle={30th USENIX security symposium (USENIX Security 21)},
  pages={2633--2650},
  year={2021}
}

@article{parikh2022canary,
  title={Canary extraction in natural language understanding models},
  author={Parikh, Rahil and Dupuy, Christophe and Gupta, Rahul},
  journal={arXiv preprint arXiv:2203.13920},
  year={2022}
}

@inproceedings{song2020information,
  title={Information leakage in embedding models},
  author={Song, Congzheng and Raghunathan, Ananth},
  booktitle={Proceedings of the 2020 ACM SIGSAC conference on computer and communications security},
  pages={377--390},
  year={2020}
}

@article{struppek2022plug,
  title={Plug \& play attacks: Towards robust and flexible model inversion attacks},
  author={Struppek, Lukas and Hintersdorf, Dominik and Correia, Antonio De Almeida and Adler, Antonia and Kersting, Kristian},
  journal={arXiv preprint arXiv:2201.12179},
  year={2022}
}

@inproceedings{zhang2020secret,
  title={The secret revealer: Generative model-inversion attacks against deep neural networks},
  author={Zhang, Yuheng and Jia, Ruoxi and Pei, Hengzhi and Wang, Wenxiao and Li, Bo and Song, Dawn},
  booktitle={Proceedings of the IEEE/CVF conference on computer vision and pattern recognition},
  pages={253--261},
  year={2020}
}

@inproceedings{carlini2023extracting,
  title={Extracting training data from diffusion models},
  author={Carlini, Nicolas and Hayes, Jamie and Nasr, Milad and Jagielski, Matthew and Sehwag, Vikash and Tramer, Florian and Balle, Borja and Ippolito, Daphne and Wallace, Eric},
  booktitle={32nd USENIX Security Symposium (USENIX Security 23)},
  pages={5253--5270},
  year={2023}
}

@inproceedings{somepalli2023diffusion,
  title={Diffusion art or digital forgery? investigating data replication in diffusion models},
  author={Somepalli, Gowthami and Singla, Vasu and Goldblum, Micah and Geiping, Jonas and Goldstein, Tom},
  booktitle={Proceedings of the IEEE/CVF conference on computer vision and pattern recognition},
  pages={6048--6058},
  year={2023}
}

@article{gundavarapu2024machine,
  title={Machine unlearning in large language models},
  author={Gundavarapu, Saaketh Koundinya and Agarwal, Shreya and Arora, Arushi and Jagadeeshaiah, Chandana Thimmalapura},
  journal={arXiv preprint arXiv:2405.15152},
  year={2024}
}

@inproceedings{brown2022does,
  title={What does it mean for a language model to preserve privacy?},
  author={Brown, Hannah and Lee, Katherine and Mireshghallah, Fatemehsadat and Shokri, Reza and Tram{\`e}r, Florian},
  booktitle={Proceedings of the 2022 ACM conference on fairness, accountability, and transparency},
  pages={2280--2292},
  year={2022}
}

@article{song2019overlearning,
  title={Overlearning reveals sensitive attributes},
  author={Song, Congzheng and Shmatikov, Vitaly},
  journal={arXiv preprint arXiv:1905.11742},
  year={2019}
}

@inproceedings{setyawan2018comparison,
  title={Comparison of multinomial naive bayes algorithm and logistic regression for intent classification in chatbot},
  author={Setyawan, Muhammad Yusril Helmi and Awangga, Rolly Maulana and Efendi, Safif Rafi},
  booktitle={2018 International Conference on Applied Engineering (ICAE)},
  pages={1--5},
  year={2018},
  organization={IEEE}
}

@inproceedings{nigam2019intent,
  title={Intent detection and slots prompt in a closed-domain chatbot},
  author={Nigam, Amber and Sahare, Prashik and Pandya, Kushagra},
  booktitle={2019 IEEE 13th international conference on semantic computing (ICSC)},
  pages={340--343},
  year={2019},
  organization={IEEE}
}

@article{agrawal2023language,
  title={Do Language Models Know When They're Hallucinating References?},
  author={Agrawal, Ayush and Suzgun, Mirac and Mackey, Lester and Kalai, Adam Tauman},
  journal={arXiv preprint arXiv:2305.18248},
  year={2023}
}

@inproceedings{huang2024position,
  title={Position: Trustllm: Trustworthiness in large language models},
  author={Huang, Yue and Sun, Lichao and Wang, Haoran and Wu, Siyuan and Zhang, Qihui and Li, Yuan and Gao, Chujie and Huang, Yixin and Lyu, Wenhan and Zhang, Yixuan and others},
  booktitle={International Conference on Machine Learning},
  pages={20166--20270},
  year={2024},
  organization={PMLR}
}

@article{morley2020initial,
  title={From what to how: an initial review of publicly available AI ethics tools, methods and research to translate principles into practices},
  author={Morley, Jessica and Floridi, Luciano and Kinsey, Libby and Elhalal, Anat},
  journal={Science and engineering ethics},
  volume={26},
  number={4},
  pages={2141--2168},
  year={2020},
  publisher={Springer}
}

@inproceedings{mitchell2019model,
  title={Model cards for model reporting},
  author={Mitchell, Margaret and Wu, Simone and Zaldivar, Andrew and Barnes, Parker and Vasserman, Lucy and Hutchinson, Ben and Spitzer, Elena and Raji, Inioluwa Deborah and Gebru, Timnit},
  booktitle={Proceedings of the conference on fairness, accountability, and transparency},
  pages={220--229},
  year={2019}
}

@article{gebru2021datasheets,
  title={Datasheets for datasets},
  author={Gebru, Timnit and Morgenstern, Jamie and Vecchione, Briana and Vaughan, Jennifer Wortman and Wallach, Hanna and Iii, Hal Daum{\'e} and Crawford, Kate},
  journal={Communications of the ACM},
  volume={64},
  number={12},
  pages={86--92},
  year={2021},
  publisher={ACM New York, NY, USA}
}

@article{wu2022survey,
  title={A survey of human-in-the-loop for machine learning},
  author={Wu, Xingjiao and Xiao, Luwei and Sun, Yixuan and Zhang, Junhang and Ma, Tianlong and He, Liang},
  journal={Future Generation Computer Systems},
  volume={135},
  pages={364--381},
  year={2022},
  publisher={Elsevier}
}

@book{yeung2021guidance,
  title={Guidance for the development of AI risk and impact assessments},
  author={Yeung, Louis Au},
  year={2021},
  publisher={Center for Long-Term Cybersecurity, University of California, Berkeley}
}

@article{guo2024bias,
  title={Bias in large language models: Origin, evaluation, and mitigation},
  author={Guo, Yufei and Guo, Muzhe and Su, Juntao and Yang, Zhou and Zhu, Mengqiu and Li, Hongfei and Qiu, Mengyang and Liu, Shuo Shuo},
  journal={arXiv preprint arXiv:2411.10915},
  year={2024}
}

@article{chen2024trustworthy,
  title={Trustworthy, responsible, and safe ai: A comprehensive architectural framework for ai safety with challenges and mitigations},
  author={Chen, Chen and Gong, Xueluan and Liu, Ziyao and Jiang, Weifeng and Goh, Si Qi and Lam, Kwok-Yan},
  journal={arXiv preprint arXiv:2408.12935},
  year={2024}
}

@article{nandkishore2024transparency,
  title={Transparency in A I decision making: A survey of explainable AI methods an d applications},
  author={Nandkishore, Patidar and Mishra, S and Jain, R and others},
  journal={Advances of Robotic Technology},
  volume={2},
  number={1},
  year={2024}
}

@article{chua2024ai,
  title={Ai safety in generative ai large language models: A survey},
  author={Chua, Jaymari and Li, Yun and Yang, Shiyi and Wang, Chen and Yao, Lina},
  journal={arXiv preprint arXiv:2407.18369},
  year={2024}
}

@article{rohde2021sustainability,
title={Sustainability challenges of artificial intelligence and policy implications},
author={Rohde, Friederike and Gossen, Maike and Wagner, Josephin and Santarius, Tilman},
journal={{\"O}kologisches Wirtschaften-Fachzeitschrift},
volume={36},
number={O1},
pages={36--40},
year={2021}
}

@article{wilson2022sustainable,
title={Sustainable AI: An integrated model to guide public sector decision-making},
author={Wilson, Christopher and Van Der Velden, Maja},
journal={Technology in Society},
volume={68},
pages={101926},
year={2022},
publisher={Elsevier}
}

@article{henderson2020towards,
title={Towards the systematic reporting of the energy and carbon footprints of machine learning},
author={Henderson, Peter and Hu, Jieru and Romoff, Joshua and Brunskill, Emma and Jurafsky, Dan and Pineau, Joelle},
journal={Journal of Machine Learning Research},
volume={21},
number={248},
pages={1--43},
year={2020}
}

@article{lacoste2019quantifying,
title={Quantifying the carbon emissions of machine learning},
author={Lacoste, Alexandre and Luccioni, Alexandra and Schmidt, Victor and Dandres, Thomas},
journal={arXiv preprint arXiv:1910.09700},
year={2019}
}

@article{patterson2021carbon,
title={Carbon emissions and large neural network training},
author={Patterson, David and Gonzalez, Joseph and Le, Quoc and Liang, Chen and Munguia, Lluis-Miquel and Rothchild, Daniel and So, David and Texier, Maud and Dean, Jeff},
journal={arXiv preprint arXiv:2104.10350},
year={2021}
}

@inproceedings{strubell2020energy,
title={Energy and policy considerations for modern deep learning research},
author={Strubell, Emma and Ganesh, Ananya and McCallum, Andrew},
booktitle={Proceedings of the AAAI conference on artificial intelligence},
volume={34},
number={09},
pages={13693--13696},
year={2020}
}

@article{luccioni2023counting,
title={Counting carbon: A survey of factors influencing the emissions of machine learning},
author={Luccioni, Alexandra Sasha and Hernandez-Garcia, Alex},
journal={arXiv preprint arXiv:2302.08476},
year={2023}
}

@article{luccioni2023estimating,
title={Estimating the carbon footprint of bloom, a 176b parameter language model},
author={Luccioni, Alexandra Sasha and Viguier, Sylvain and Ligozat, Anne-Laure},
journal={Journal of Machine Learning Research},
volume={24},
number={253},
pages={1--15},
year={2023}
}

@inproceedings{chien2023reducing,
title={Reducing the Carbon Impact of Generative AI Inference (today and in 2035)},
author={Chien, Andrew A and Lin, Liuzixuan and Nguyen, Hai and Rao, Varsha and Sharma, Tristan and Wijayawardana, Rajini},
booktitle={Proceedings of the 2nd workshop on sustainable computer systems},
pages={1--7},
year={2023}
}

@article{li2024toward,
title={Toward sustainable genai using generation directives for carbon-friendly large language model inference},
author={Li, Baolin and Jiang, Yankai and Gadepally, Vijay and Tiwari, Devesh},
journal={arXiv preprint arXiv:2403.12900},
year={2024}
}

@inproceedings{luccioni2024power,
title={Power hungry processing: Watts driving the cost of ai deployment?},
author={Luccioni, Sasha and Jernite, Yacine and Strubell, Emma},
booktitle={Proceedings of the 2024 ACM conference on fairness, accountability, and transparency},
pages={85--99},
year={2024}
}

@article{kamruzzaman2023efficient,
title={Efficient Sentiment Analysis: A Resource-Aware Evaluation of Feature Extraction Techniques, Ensembling, and Deep Learning Models},
author={Kamruzzaman, Mahammed and Kim, Gene Louis},
journal={arXiv preprint arXiv:2308.02022},
year={2023}
}

@inproceedings{jalilifard2021semantic,
title={Semantic sensitive TF-IDF to determine word relevance in documents},
author={Jalilifard, Amir and Carid{\'a}, Vinicius Fernandes and Mansano, Alex Fernandes and Cristo, Rogers S and da Fonseca, Felipe Penhorate Carvalho},
booktitle={Advances in Computing and Network Communications: Proceedings of CoCoNet 2020, Volume 2},
pages={327--337},
year={2021},
organization={Springer}
}

@article{reinanda2020knowledge,
title={Knowledge graphs: An information retrieval perspective},
author={Reinanda, Ridho and Meij, Edgar and de Rijke, Maarten and others},
journal={Foundations and Trends{\textregistered} in Information Retrieval},
volume={14},
number={4},
pages={289--444},
year={2020},
publisher={Now Publishers, Inc.}
}

@misc{unesco_aiethics,
title = {Ethics of Artificial Intelligence},
year={2021},
author={UNESCO},
howpublished = {\url{https://www.unesco.org/en/artificial-intelligence/recommendation-ethics}},
note = {Accessed: 2025-04-21}
}

@misc{oecd2024principles,
title = {OECD AI Principles overview},
year={2024},
author={OECD},
howpublished = {\url{https://oecd.ai/en/ai-principles}},
note = {Accessed: 2025-03-24}
}

@article{chatila2019ieee,
title={The IEEE global initiative on ethics of autonomous and intelligent systems},
author={Chatila, Raja and Havens, John C},
journal={Robotics and well-being},
pages={11--16},
year={2019},
publisher={Springer}
}

@article{ai2019high,
  title={High-level expert group on artificial intelligence},
  author={EU-HLEG},
  journal={Ethics guidelines for trustworthy AI},
  volume={6},
  year={2019},
  publisher={European Commission. Available at: https://ec. europa. eu/digital-single~…}
}

@misc{eu2024regulation,
title = {The EU Artificial Intelligence Act},
year={2024},
author={EU-AI-Act},
howpublished = {\url{https://eur-lex.europa.eu/legal-content/EN/TXT/HTML/?uri=OJ:L_202401689}},
note = {Accessed: 2025-02-10}
}

@misc{iso42001,
  title        = {ISO/IEC 42001:2023 — Information technology — Artificial intelligence — Management system},
  author       = {{ISO/IEC}},
  year         = {2023},
  howpublished = {\url{https://www.iso.org/standard/81230.html}},
  note         = {Accessed: 2025-04-25}
}

@article{herrera2025overview,
  title={An overview of model uncertainty and variability in LLM-based sentiment analysis. Challenges, mitigation strategies and the role of explainability},
  author={Herrera-Poyatos, David and Pel{\'a}ez-Gonz{\'a}lez, Carlos and Zuheros, Cristina and Herrera-Poyatos, Andr{\'e}s and Tejedor, Virilo and Herrera, Francisco and Montes, Rosana},
  journal={arXiv preprint arXiv:2504.04462},
  year={2025}
}

@inproceedings{scherzinger2019best,
  title={The best of both worlds: Challenges in linking provenance and explainability in distributed machine learning},
  author={Scherzinger, Stefanie and Seifert, Christin and Wiese, Lena},
  booktitle={2019 IEEE 39th International Conference on Distributed Computing Systems (ICDCS)},
  pages={1620--1629},
  year={2019},
  organization={IEEE}
}

@article{herrera2025responsible,
  title={Responsible Artificial Intelligence Systems: A Roadmap to Society's Trust through Trustworthy AI, Auditability, Accountability, and Governance},
  author={Herrera-Poyatos, Andr{\'e}s and Del Ser, Javier and de Prado, Marcos L{\'o}pez and Wang, Fei-Yue and Herrera-Viedma, Enrique and Herrera, Francisco},
  journal={arXiv preprint arXiv:2503.04739},
  year={2025}
}

@article{freeman2024exploring,
  title={Exploring Memorization and Copyright Violation in Frontier LLMs: A Study of the New York Times v. OpenAI 2023 Lawsuit},
  author={Freeman, Joshua and Rippe, Chloe and Debenedetti, Edoardo and Andriushchenko, Maksym},
  journal={arXiv preprint arXiv:2412.06370},
  year={2024}
}

@article{sag2023copyright,
  title={Copyright safety for generative AI},
  author={Sag, Matthew},
  journal={Hous. L. Rev.},
  volume={61},
  pages={295},
  year={2023},
  publisher={HeinOnline}
}

@article{le2023problem,
  title={The problem with annotation. Human labour and outsourcing between France and Madagascar},
  author={Le Ludec, Cl{\'e}ment and Cornet, Maxime and Casilli, Antonio A},
  journal={Big Data \& Society},
  volume={10},
  number={2},
  pages={20539517231188723},
  year={2023},
  publisher={SAGE Publications Sage UK: London, England}
}

@inproceedings{sun2023human,
  title={Human-in-the-loop interaction for continuously improving generative model in conversational agent for behavioral intervention},
  author={Sun, Xin and Bosch, Jos A and De Wit, Jan and Krahmer, Emiel},
  booktitle={Companion Proceedings of the 28th International Conference on Intelligent User Interfaces},
  pages={99--101},
  year={2023}
}

@article{achiam2023gpt,
  title={Gpt-4 technical report},
  author={Achiam, Josh and Adler, Steven and Agarwal, Sandhini and Ahmad, Lama and Akkaya, Ilge and Aleman, Florencia Leoni and Almeida, Diogo and Altenschmidt, Janko and Altman, Sam and Anadkat, Shyamal and others},
  journal={arXiv preprint arXiv:2303.08774},
  year={2023}
}

@article{tang2023verifai,
title={VerifAI: verified generative AI},
author={Tang, Nan and Yang, Chenyu and Fan, Ju and Cao, Lei and Luo, Yuyu and Halevy, Alon},
journal={arXiv preprint arXiv:2307.02796},
year={2023}
}

@article{floridi2022capai,
title={CapAI-A procedure for conducting conformity assessment of AI systems in line with the EU artificial intelligence act},
author={Floridi, Luciano and Holweg, Matthias and Taddeo, Mariarosaria and Amaya, Javier and M{\"o}kander, Jakob and Wen, Yuni},
journal={Available at SSRN 4064091},
year={2022}
}

@article{carrad2022australian,
title={Australian local government policies on creating a healthy, sustainable, and equitable food system: analysis in New South Wales and Victoria},
author={Carrad, Amy and Aguirre-Bielschowsky, Ikerne and Reeve, Belinda and Rose, Nick and Charlton, Karen},
journal={Australian and New Zealand Journal of Public Health},
volume={46},
number={3},
pages={332--339},
year={2022},
publisher={Elsevier}
}

@article{sutton2025understanding,
title={Understanding innovation in the context of local economic development: An analysis of cities’ innovation-based policies in Ontario, Canada},
author={Sutton, Jesse and Phan, Selina and Arku, Godwin and Hutchenreuther, John and Cleave, Evan},
journal={Local Development \& Society},
volume={6},
number={1},
pages={97--115},
year={2025},
publisher={Taylor \& Francis}
}

@inproceedings{xia2024towards,
title={Towards a responsible ai metrics catalogue: A collection of metrics for ai accountability},
author={Xia, Boming and Lu, Qinghua and Zhu, Liming and Lee, Sung Une and Liu, Yue and Xing, Zhenchang},
booktitle={Proceedings of the IEEE/ACM 3rd International Conference on AI Engineering-Software Engineering for AI},
pages={100--111},
year={2024}
}

@article{matthews2020patterns,
title={Patterns and anti-patterns, principles and pitfalls: accountability and transparency in AI},
author={Matthews, Jeanna},
journal={AI Magazine},
volume={41},
number={1},
pages={82--89},
year={2020}
}

@article{smith2021clinical,
title={Clinical AI: opacity, accountability, responsibility and liability},
author={Smith, Helen},
journal={Ai \& Society},
volume={36},
number={2},
pages={535--545},
year={2021},
publisher={Springer}
}

@article{blackman2022you,
title={WHY YOU NEED AN AI ETHICS COMMITTEE Expert oversight will help you safe-guard your data and your brand},
author={Blackman, Reid},
journal={Harvard Business Review},
volume={100},
number={7-8},
pages={118--125},
year={2022},
publisher={HARVARD BUSINESS SCHOOL PUBLISHING CORPORATION 300 NORTH BEACON STREET~…}
}

@article{azaria2024chatgpt,
  title={ChatGPT is a remarkable tool—for experts},
  author={Azaria, Amos and Azoulay, Rina and Reches, Shulamit},
  journal={Data Intelligence},
  volume={6},
  number={1},
  pages={240--296},
  year={2024},
  publisher={MIT Press One Rogers Street, Cambridge, MA 02142-1209, USA journals-info~…}
}

@inproceedings{ladhak2023pre,
  title={When do pre-training biases propagate to downstream tasks? a case study in text summarization},
  author={Ladhak, Faisal and Durmus, Esin and Suzgun, Mirac and Zhang, Tianyi and Jurafsky, Dan and McKeown, Kathleen and Hashimoto, Tatsunori B},
  booktitle={Proceedings of the 17th Conference of the European Chapter of the Association for Computational Linguistics},
  pages={3206--3219},
  year={2023}
}

@article{steen2023bias,
  title={Bias in News Summarization: Measures, Pitfalls and Corpora},
  author={Steen, Julius and Markert, Katja},
  journal={arXiv preprint arXiv:2309.08047},
  year={2023}
}

@article{brown2023fair,
  title={How (un) fair is text summarization?},
  author={Brown, Hannah and Shokri, Reza},
  year={2023}
}

@article{lee2022neus,
  title={NeuS: neutral multi-news summarization for mitigating framing bias},
  author={Lee, Nayeon and Bang, Yejin and Yu, Tiezheng and Madotto, Andrea and Fung, Pascale},
  journal={arXiv preprint arXiv:2204.04902},
  year={2022}
}
